\documentclass[journal,twocolumn,10pt]{IEEEtran}
\usepackage{amsmath,epsfig} 
\usepackage{lipsum}

\usepackage{floatflt} 
\usepackage{paralist} 
\usepackage{algorithm} 
\usepackage{graphicx}
\usepackage{amsthm}
\usepackage{amssymb}
\usepackage{listings}    
\usepackage{xcolor}
\usepackage{hyperref}
\usepackage{indentfirst}
\usepackage{booktabs}
\usepackage{subfigure}
\usepackage{caption}
\usepackage{calligra}
\usepackage{bm}
\usepackage{fancyhdr}
\usepackage{float}
\usepackage{pifont}
\usepackage{colortbl} 
\usepackage{amsmath}
\usepackage{MnSymbol}%
\usepackage{amsbsy}
\usepackage{soul} 

\def\a{\mathbf{a}}
\def\b{\mathbf{b}}

\def\x{\mathbf{x}}

\def\y{\mathbf{y}}

\def\e{\mathbf{e}}

\def\h{\mathbf{h}}

\def\n{\mathbf{n}}
\def\oo{\mathbf{o}}
\def\p{\mathbf{p}}

\def\z{\mathbf{z}}
\def\S{\mathcal{S} }

\def\N{\mathcal{N}}
\def\V{\mathcal{V}}
\def\E{\mathcal{E}}
\def\M{\mathcal{M}}

\newcommand{\mypar}[1]{{\bf #1.}}
\theoremstyle{definition}

\DeclareMathOperator{\Adj}{A}

\DeclareMathOperator{\D}{D}

\DeclareMathOperator{\X}{X}
\DeclareMathOperator{\Yy}{Y}
\DeclareMathOperator{\Hh}{H}

\DeclareMathOperator{\Z}{Z}

\newcommand{\R}{\ensuremath{\mathbb{R}}}

\DeclareMathOperator{\Id}{I}

\title{3D Point Cloud Processing and Learning for Autonomous Driving}

\author{Siheng~Chen, Baoan Liu, Chen Feng, Carlos Vallespi-Gonzalez, Carl Wellington
\thanks{S. Chen is with Mitsubishi Electric Research Laboratories, Cambridge, MA, USA. Email: schen@merl.com. B. Liu is with Precivision Technologies, Inc., Pittsburgh, PA, USA. Email: baoanliu@precivision.tech.  C. Feng is with New York University, Brooklyn, NY, USA. Email: cfeng@nyu.edu. 
C. Vallespi-Gonzalez and C. Wellington are with Uber Advanced Technologies Group, Pittsburgh, PA, USA. Emails: cvallespi@uber.com, cwellington@uber.com.  }
}

\begin{document}

\maketitle
\tableofcontents

\begin{abstract}
We present a review of 3D point cloud processing and learning for autonomous driving. As one of the most important sensors in autonomous vehicles, light detection and ranging (LiDAR) sensors collect 3D point clouds that precisely record the external surfaces of objects and scenes. The tools for 3D point cloud processing and learning are critical to the map creation, localization, and perception modules in an autonomous vehicle. While much attention has been paid to data collected from cameras, such as images and videos, an increasing number of researchers have recognized the importance and significance of LiDAR in autonomous driving and have proposed processing and learning algorithms to exploit 3D point clouds. We review the recent progress in this research area and summarize what has been tried and what is needed for practical and safe autonomous vehicles. We also offer perspectives on open issues that are needed to be solved in the future.
\end{abstract}

\section{Introduction and Motivation}
\label{sec:intro}

\subsection{Autonomous driving: Significance, history and current state}
\label{sec:intro_autonomous_driving}
As one of the most exciting engineering projects of the modern world, autonomous driving is an aspiration for many researchers and engineers across generations. It is a goal that might fundamentally redefine the future of human society and everyone's daily life. Once autonomous driving becomes mature, we will witness a transformation of public transportation, infrastructure and the appearance of our cities. The world is looking forward to exploiting autonomous driving to reduce traffic accidents caused by driver errors, to save drivers' time and liberate the workforce, as well as to save parking spaces, especially in the urban area~\cite{TaeihaghL:19}.

It has taken decades of effort to get closer to the goal of autonomous driving. From the 1980s through the DARPA Grand Challenge in 2004 and the DARPA Urban Challenge in 2007, the research on autonomous driving was primarily conducted in the U.S. and Europe, yielding incremental progresses in driving competence in various situations~\cite{NRC:02}. In 2009, Google started a research project on self-driving cars, and later created Waymo to commercialize the accomplishment based on their early technical success. Around 2013-2014, the rise of deep neural networks brought on the revolution of practical computer vision and machine learning. This emergence made people believe that many technical bottlenecks of autonomous driving could be fundamentally solved. In 2015, Uber created the Uber Advanced Technologies Group with the aim to enable autonomous vehicles to complete scalable ride-sharing services. This aim has become a common deployment strategy within the industry. Currently, there are numerous high-tech companies, automobile manufacturers, and start-up companies working on autonomous-driving technologies, including Apple, Aptiv, Argo AI, Aurora, Baidu, GM Cruise, Didi, Lyft, Pony.ai, Tesla, Zoox, the major automobile companies, and many others~\cite{Badue:19}. These companies have ambitious goals to achieve SAE level 4\footnote{SAE International, a transportation standards organization, introduced the J3016 standard, which defines six levels of driving automation; See details in~\url{https://www.sae.org/news/2019/01/sae-updates-j3016-automated-driving-graphic}. It ranges from SAE Level Zero (no automation) to SAE Level 5 (full automation). One turning point occurs between Levels 2 and 3, where the driving responsibility shifts from a human driver to an autonomous system, and another turning point occurs between Levels 3 and 4, where the human no longer drives under any circumstances.} in the near future. Although there has been significant progress across many groups in industry and academia, there is still much work to be done. The efforts from both industry and academia are needed to achieve autonomous driving. Recently, there have been many discussions and hypotheses about the progress and the future of autonomous driving; however, few thoughts from those who push industrial-level self-driving technologies from the frontline are publicly accessible. In this article, we provide a unifying perspective from both practitioners and researchers.

\begin{figure*}[t]
  \begin{center}
	\includegraphics[width=1.9\columnwidth]{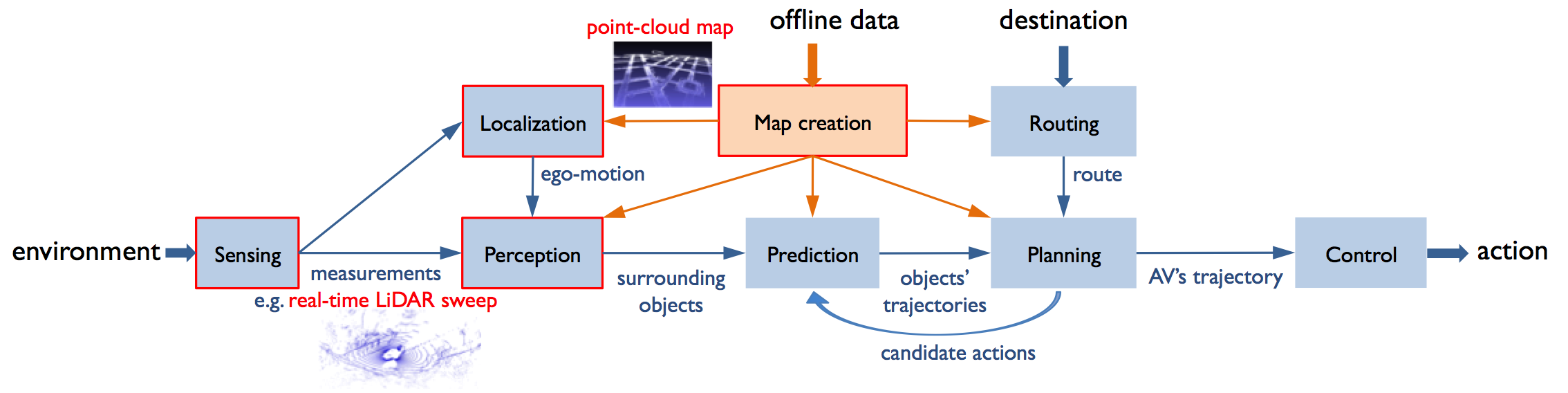}     
\end{center}

\caption{\label{fig:autonomous_system} High-level block diagram of a typical autonomous system. A high-definition map is built offline. At runtime, the online system is given a destination. The system then senses its environment, localizes itself to the map, perceives the world around it and makes corresponding predictions of future motion for these objects. The motion planner uses these predictions to plan a safe trajectory for an autonomous vehicle (AV) to follow the route to the destination that is executed by the controller. Note that two types of 3D point clouds are used in this autonomous system: a point-cloud map, created by the map creation module and consumed by the localization module, and a real-time LiDAR sweep, collected by the sensing module and consumed by the localization and perception modules.}
\end{figure*}

In industry, an autonomous system usually includes a series of modules with complicated internal dependencies. Most modules are still far from being perfect due to a number of technical bottlenecks and the long-tail issues~\cite{BansalKO:18}. Additionally, a small error from one module can cause problems in subsequent modules and potentially result in a substantial failure at the system level. There has been some initial research on end-to-end systems where the entire system is trained end-to-end and information can flow from sensors directly to the final motion-planning or control decisions. These systems offer the promise to reduce internal dependency challenges; however, these systems often lack explainability and are difficult to analyze. Although significant progress has been made, there remain many open challenges in designing a practical autonomous system that can achieve the goal of full self-driving.

\subsection{A tour of an autonomous system}
\label{sec:intro_autonomous_system}
An autonomous system typically includes the sensing, map creation, localization,  perception,  prediction,  routing,  motion-planning, and control modules~\cite{UrmsonABBBCDDGGGHHHKKLMMPPRRSSSSSWWZBBDLNSZSTDF:09}; see Figure~\ref{fig:autonomous_system}. A high-definition map is created offline. At runtime, the online system is given a destination. The system then senses its environment, localizes itself to the map, perceives the world around it and makes corresponding predictions of future motion for these objects. The motion planner uses these predictions to plan a safe trajectory for an autonomous vehicle (AV) to follow the route to the destination that is executed by the controller.

\mypar{Sensing module} 
To ensure reliability, autonomous driving usually requires multiple types of sensors. Cameras, radio detection and ranging (RADAR), light detection and ranging (LiDAR) and ultrasonic sensors are  most commonly used. Among those sensors, LiDAR is particularly interesting because it directly provides a precise 3D representation of a scene. Although the techniques for 3D reconstruction and depth estimation based on 2D images have been significantly improved with the development of deep learning based computer vision algorithms, the resulting estimations are still not always precise or reliable. Besides algorithmic constraints, fundamental bottlenecks also include inherent exponential range error growth in depth estimation, poor performance in low light, and the high computational cost of processing high-resolution images. On the other hand, LiDAR measures 3D information through direct physical sensing. A real-time LiDAR sweep consists of a large number of 3D points; called a 3D point cloud\footnote{The measurements from RADAR and Ultrasound are also called 3D point clouds, but we focus on 3D point clouds collected by LiDAR.}. Each 3D point records the range from the LiDAR to an object's external surface, which can be transformed into the precise 3D coordinate. These 3D point clouds are extremely valuable for an autonomous vehicle to localize itself and detect surrounding objects in the 3D world.~\emph{The vast majority of companies and researchers rely heavily on LiDAR to build a reliable autonomous vehicle~\cite{MeyerLKVW:19}.} This is why we believe that advanced techniques for 3D point cloud processing and learning are indispensable for autonomous driving. 

\mypar{Map creation module}
Map creation is the task of creating a high-definition (HD) map, which is a precise heterogeneous map representation of the static 3D environment and traffic rules.
A HD map usually contains two map layers: a~\emph{point-cloud map}, representing 3D geometric information of surroundings, and a ~\emph{traffic-rule-related semantic feature map}, containing road boundaries, traffic lanes, traffic signs, traffic lights, etc. These two map layers are aligned together in the 3D space and provide detailed navigation information. As one map layer, the point-cloud map is a dense 3D point cloud and mainly used for providing localization prior. Different from common maps designed for humans, an HD map is designed for autonomous vehicles. The map creation module is crucial because an HD map provides valuable prior environmental information; see details in Section~\ref{sec:map}.

\mypar{Localization module} Localization is the task of finding the ego-position of an autonomous vehicle relative to a reference position in the HD map. This module is crucial because an autonomous vehicle must localize itself in order to use the correct lane and other important priors in the HD map. One of the core techniques is 3D point cloud registration; that is, estimating the precise location of an autonomous vehicle by matching real-time LiDAR sweeps to the offline HD map; see details in Section~\ref{sec:localization}.

\mypar{Perception}
Perception is the task of perceiving the surrounding environment and extracting information that is related to navigation. This module is crucial because the perception module is the visual system of an autonomous vehicle, which should detect, track and classify objects in the 3D scene. It used to be considered as the technical bottleneck of autonomous driving. Recently, with large-scale training data and developments of advanced machine learning algorithms, the overall performance of the perception module has achieved tremendous improvement. Some core techniques include 2D object detection and 3D object detection. 2D object detection becomes relatively mature, while 3D object detection is based on real-time LiDAR sweeps and becomes an increasingly hot research topic; see details in Section~\ref{sec:perception}.

\mypar{Prediction} Prediction is the task of predicting the future potential trajectories of each object in the 3D scene. This module is crucial because an autonomous vehicle needs to know the possible future behaviors of nearby objects to plan a safe trajectory.

\mypar{Routing} Routing is the task of designing a high-level path from the starting position to the destination for an autonomous vehicle. The output of this module provides a high-level guideline for the motion-planning module.

\mypar{Planning} Motion planning is the task of designing a trajectory for an autonomous vehicle based on the state of current cars, surrounding environment and the destination. This module is crucial because an autonomous vehicle needs to know how to react to the surrounding environment.

\mypar{Control} Control is the task of executing the commands from the motion-planning module. It takes charge of controlling the actuators of the steering wheel, throttle, and brakes.

\subsection{Overview of 3D point cloud processing and learning}
\label{sec:intro_PCPL}
As mentioned earlier, LiDAR provides indispensable 3D information for autonomous driving. We now move on to the processing and learning techniques that convert raw measurements into useful information.

\mypar{Usages in autonomous driving}
\label{sec:pc_usages}
Two types of 3D point clouds are commonly used in an autonomous vehicle: a real-time LiDAR sweep and a point-cloud map, which is one layer in the HD map; see Figure~\ref{fig:autonomous_system}.  A point-cloud map provides prior environmental information: the localization module uses a point-cloud map as a reference in 3D point cloud registration to determine the position of the autonomous vehicle, and the perception module uses a point-cloud map to help split the foreground and the background. On the other hand, real-time LiDAR sweeps are consumed by the localization module to register against the point-cloud map, and by the perception module to detect surrounding objects in the 3D scene.  Therefore, 3D point cloud processing and learning are critical to build the map creation, localization and perception modules in an autonomous system.

\mypar{Recent progress in academia}
Sensors capture data and data feeds algorithms. During the development of RADAR, acoustic sensors and communication systems, 1D signal processing experienced a rapid growth during the past century, leading to a revolutionary impact on digital communication systems. With the popularization of cameras and televisions, 2D image processing experienced a rapid growth during the past 30 years, resulting in a significant change to photography, entertainment, and surveillance.  With the increasing needs from industrial robotics, autonomous driving and augmented reality, 3D sensing techniques is experiencing rapid development recently. At the same time, the algorithms to process and learn from 3D point clouds are starting to get much attention in academia. The following discussion is divided into two parts: 3D point cloud processing, which handles  3D point clouds from a signal-processing perspective, and 3D point cloud learning,  which handles  3D point clouds  from a machine-learning perspective.

\mypar{3D point cloud processing} 
3D point cloud processing is the process of analyzing and modifying a 3D point cloud to optimize its transmission, storage and quality through various mathematical and computational algorithms. Even though the processing algorithms could be significantly different, many processing tasks are naturally extended from 1D signal processing and 2D image processing. For example, 3D point cloud compression is the 3D counterpart of image compression that aims to reduce the cost for storage or transmission of a 3D point cloud; 3D point cloud denoising is the 3D counterpart of image denoising that aims to remove noise from a 3D point cloud; 3D point cloud registration is the 3D counterpart of image registration that aims to align two or more 3D point clouds of the same scene; and 3D point cloud downsampling and upsampling are the 3D counterpart of image scaling that aims to change the resolution (number of points) in a 3D point cloud.

\mypar{3D point cloud learning}
3D point cloud learning is the process of interpreting and understanding a 3D point cloud. With the powerful tools of deep neural networks, computer vision researchers aim to extend the success from images and videos to 3D point clouds. Two primary learning problems are 3D point cloud recognition and segmentation. Similarly to the cases for 2D images, 3D point cloud recognition aims to classify a given 3D point cloud into a predefined class category and  3D point cloud segmentation aims to partition a given 3D point cloud into multiple segments. Due to the irregular format of 3D point clouds, one of the biggest challenges for designing a learning algorithm is to formulate efficient data structures to represent 3D point clouds. Some algorithms transform 3D point clouds to regular 3D voxels, so that 3D convolutions can be used for the analysis; however, they have to make a trade-off between resolution and memory.  To handle raw point clouds directly, PointNet~\cite{QiSMG:17} uses point-wise multilayer perceptrons (MLPs) and max-pooling to ensure the permutation invariance. After that, a series of 3D deep learning methods follow PointNet as their base networks.

\mypar{Relations between academia and industry} 
The technical transition from 1D time-series to 2D images is quite natural, because both types of data are supported on regular-spacing structures; however, the technical transition from 2D images to 3D point clouds is not straightforward because those points are irregularly scattered in a 3D space. Numerous popular methods to handle 3D point clouds are proposed heuristically by practitioners. Therefore, there is a substantial room for both researchers and practitioners to collaborate and solve fundamental tasks on 3D point cloud processing and learning, so that we can accelerate the progress of autonomous driving. 

\subsection{Outline}
\label{sec:outline}
The outline of this article is as follows: Section~\ref{sec:ingredients} presents key ingredients of 3D point cloud processing and learning. It starts by explaining common properties of a 3D point cloud, followed by various approaches to represent a 3D point cloud. It then presents modern methods to process and learn from a 3D point cloud. Sections~\ref{sec:map},~\ref{sec:localization}, and~\ref{sec:perception} cover the state-of-the-art methods and challenges about
3D point cloud processing and learning in the map creation, localization and perception modules of an autonomous system, respectively. We specifically consider these three modules because they heavily rely on 3D point clouds to achieve reliable performance. In each module, we discuss~\emph{what} this module is specifically working on; ~\emph{why} 3D point cloud processing and learning are significant for this module; and~\emph{how} 3D point cloud processing and learning make a difference in this module. Section~\ref{sec:perspectives} concludes with discussion and pointers to future directions. In Appendix, we compare the perspectives between academia and industry in Section I, illustrate the latest qualitative results in Section II, and overview a series of elementary tasks about 3D point clouds that have received much attention in academia in Section III.

\section{Key Ingredients of 3D Point Cloud Processing and Learning}
\label{sec:ingredients}
In this section, we introduce basic tools of 3D point cloud processing and learning. We start with the key properties of 3D point clouds. We next evaluate some options for representing a 3D point cloud. Finally, we review a series of popular tools to handle 3D point clouds. Those tools have received great attention in academia. Even some of them might not be directly applied to an autonomous system, it is still worth mentioning because they could inspire new techniques, which are potentially useful to autonomous driving.

\subsection{Properties}
\label{sec:properties}
As discussed in Section~\ref{sec:pc_usages}, we consider two typical types of 3D point clouds in autonomous driving: real-time LiDAR sweeps and point-cloud maps.

\mypar{Real-time LiDAR sweeps}
Because of the sensing mechanism, for each 3D point in a real-time LiDAR sweep, we can trace its associated laser beam and captured time stamp. One real-time LiDAR sweep can naturally be organized on a 2D image, whose  $x$-axis is the time stamp and $y$-axis is the laser ID. We thus consider each individual real-time LiDAR sweep as an~\emph{organized 3D point cloud}. For example, a Velodyne HDL-64E has 64 separate lasers and each laser fires thousands of times per second to capture a 360-degree field of view. We thus obtain a set of 3D points associated with $64$ elevation angles and thousands of azimuth angles\footnote{In a real-time LiDAR sweep, the vertical resolution is usually much lower than the horizontal resolution.}. Each collected 3D point is associated with a range measurement, an intensity value and a high precision GPS time stamps. Note that for a global-shutter image, the pixel values are collected by a charge-coupled device (CCD) at the same time; however, for a real-time LiDAR sweep, the 3D points are collected at various time stamps. For the same laser, firings happen sequentially to collect 3D points; for different lasers, firings are not synchronized either; thus, the collected 3D points are not perfectly aligned on a 2D regular lattice.  Since the arrangement of 64 lasers follows a regular angular spacing, the point density of a real-time LiDAR sweep changes over the range; that is, we collect many more 3D points from nearby objects than from far-away objects. Moreover, a real-time LiDAR sweep naturally suffers from the occlusion; that is, we get 3D points only from the sides of objects facing the LiDAR. To summarize, some key properties of a real-time LiDAR sweep include:
\begin{itemize}
\item Pseudo 3D. A real-time LiDAR sweep arranges 3D points approximately on a 2D lattice. Due to the non-perfect synchronization, 3D points are not perfectly aligned on a 2D lattice. Meanwhile, unlike a 3D point cloud obtained from multiple views, a real-time LiDAR sweep only reflects a specific view; we thus consider its dimension~\emph{pseudo 3D};

\item Occlusion. Each individual real-time LiDAR sweep records the 3D environment almost from a single viewpoint\footnote{Since the autonomous vehicle could move in real-time, the viewpoint of LiDAR would change slightly.}. A front object would occlude the other objects behind it; and

\item Sparse point clouds. Compared to a 2D image, a real-time LiDAR sweep is usually sparse representations of objects, especially for far-away objects. It cannot provide detailed 3D shape information of objects.
\end{itemize}

\mypar{Point-cloud maps}
To create a point-cloud map, one needs to aggregate real-time LiDAR sweeps scanned from multiple autonomous vehicles across time. Since there is no straightforward way to organize a point-cloud map,
we consider it as an~\emph{unorganized 3D point cloud}. For example, for a $200 \times 200$ square meter portion of an HD map, one needs to aggregate the LiDAR sweeps around that area for 5-10 trials, leading to over $10$ millions 3D points. Since LiDAR sweeps could be collected from significantly different views, an HD map after aggregation gets denser and presents a detailed 3D shape information. To summarize, some key properties of a point-cloud map include:
\begin{itemize}
\item Full 3D. A point-cloud map aggregates multiple LiDAR sweeps from various views, which is similar to 3D data collected by scanning an object on a turntable. A point-cloud map captures information on more objects' surfaces, providing a denser and more detailed 3D representation;

\item Irregularity. 3D points in a point-cloud map are irregularly scattered in the 3D space. They come from multiple LiDAR sweeps and lose the laser ID association, causing an unorganized 3D point cloud;

\item No occlusion. A point-cloud map is an aggregation of 3D points collected from multiple viewpoints. It depicts the static 3D scene with much less occlusion;

\item Dense point clouds. A point-cloud map provides a dense point cloud, which contains detailed 3D shape information, such as high-resolution shapes and the surface normals; and

\item Semantic meanings. As another layer in the HD map, a traffic-rule-related semantic feature map contain the semantic labels of a 3D scene, including road surfaces, buildings  and trees.  Since a traffic-rule-related semantic feature map and a point-cloud map are aligned in the 3D space, we can trace the semantic meaning of each 3D point. For example, 3D points labeled as trees in a point-cloud map would help improve perception as LiDAR points on leaves of trees are usually noisy and difficult to be recognized.
\end{itemize}

\subsection{Matrix representations}
\label{sec:representations}
Representations have always been at the heart of most signal processing and machine learning techniques. A good representation lays the foundation to uncover hidden patterns and structures within data and is beneficial for subsequent tasks. A general representation of a 3D point cloud is through a set, which ignores any order of 3D points. Let $\S = \{ (\p_i, \a_i) \}_{i=1}^N$ be a set of $N$ 3D points, whose $i$th element $\p_i = [x_i, y_i, z_i] \in \R^3$ represents the 3D coordinate of the $i$th point and $\a_i$ represents other attributes of the $i$th point. A real-time LiDAR sweep usually includes the intensity $\a_i = r_i \in \R$ and a point-cloud map usually includes surface normals $\n_i \in \R^3$; thus, $\a_i = [r_i, \n_i ] \in \R^4$. For generality, we consider the feature of the $i$th point as $\x_i = (\p_i, \a_i) \in \R^d$.

For efficient storage and scientific computation, a matrix (or tensor) representation is appealing. Let $f$ be the mapping from a set of 3D points $\S$ to a matrix (or tensor) $\X$ with a pending shape. A matrix representation of a 3D point cloud is thus
$
\X \ = \  f(\S).
$
We next discuss a few typical approaches to implement the mapping $f(\cdot)$. 

\mypar{Raw points}
The most straightforward matrix representation of a 3D point cloud is to list each 3D point in the set $\S$ as one row in the matrix. Consider
\begin{eqnarray}
  \label{eq:raw_point_representation}
  \X^{\rm (raw)} \ = \  \begin{bmatrix} \x_1 & \x_2 & \cdots &
    \x_N
  \end{bmatrix}^T \ \in \ \R^{N \times d},
\end{eqnarray}
whose $i$th row $\X^{\rm (raw)}_i = \x_i  \in \R^d$
is the features of $i$th point in the 3D point cloud. 

The advantages of the raw-point-based representation are that i) it is simple and general; ii) it preserves all the information in the original set of 3D points; however, the shortcoming is that it does not explore any geometric property of 3D points. This representation is generally used in the map and the localization module of an autonomous system, where high precision is needed. 

\begin{figure*}[thb]
  \begin{center}
    \begin{tabular}{ccc}
   \includegraphics[width=0.6\columnwidth]{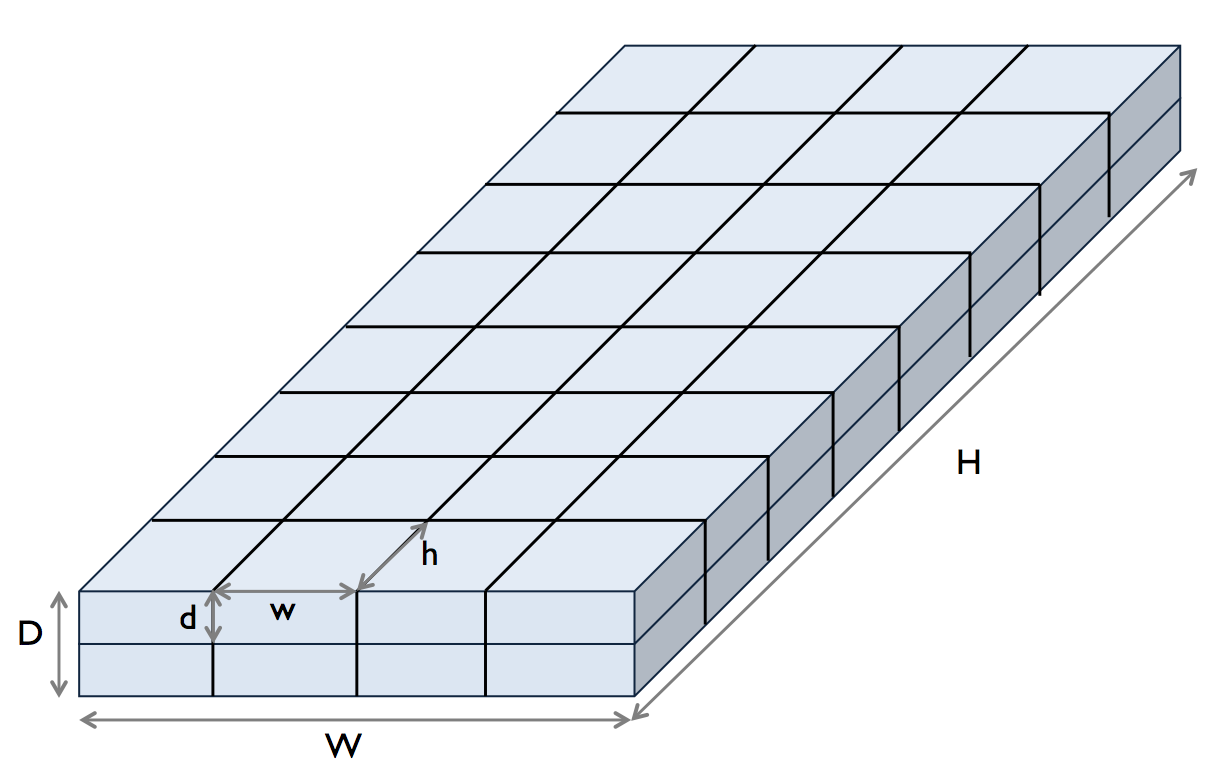}     & \includegraphics[width=0.6\columnwidth]{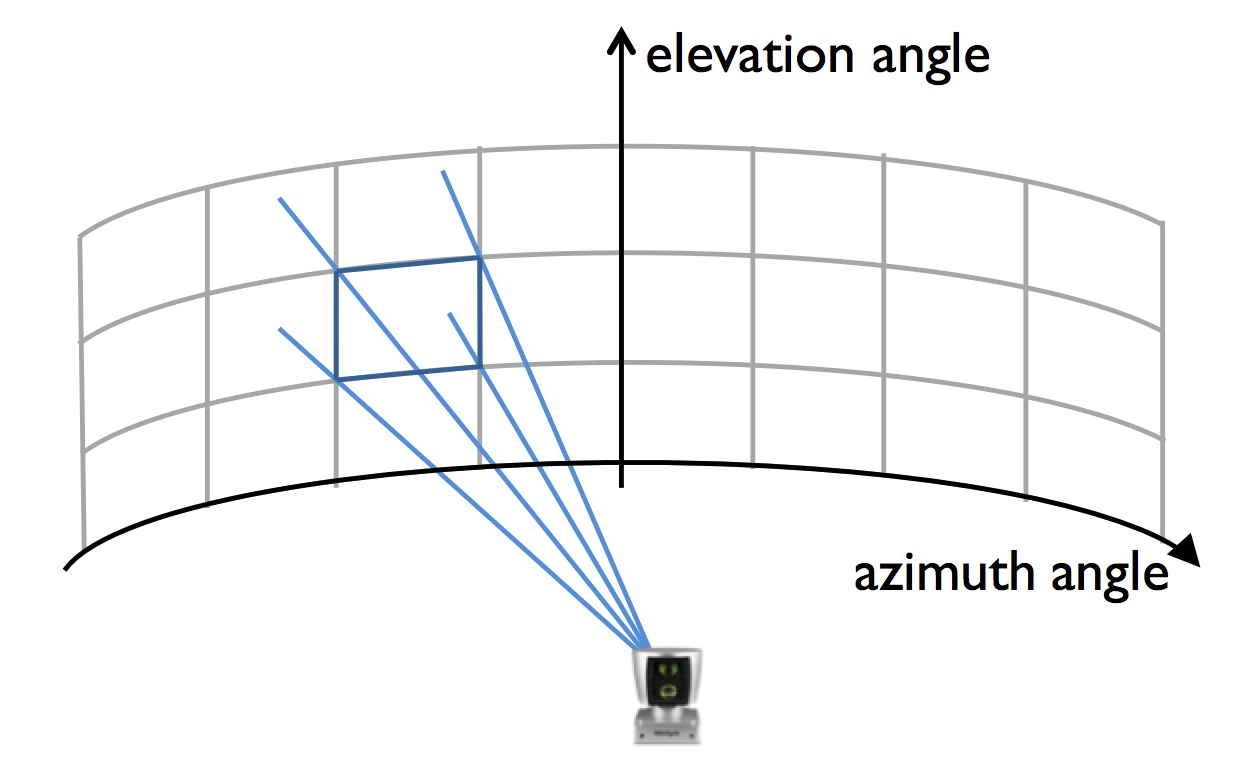}    & \includegraphics[width=0.4\columnwidth]{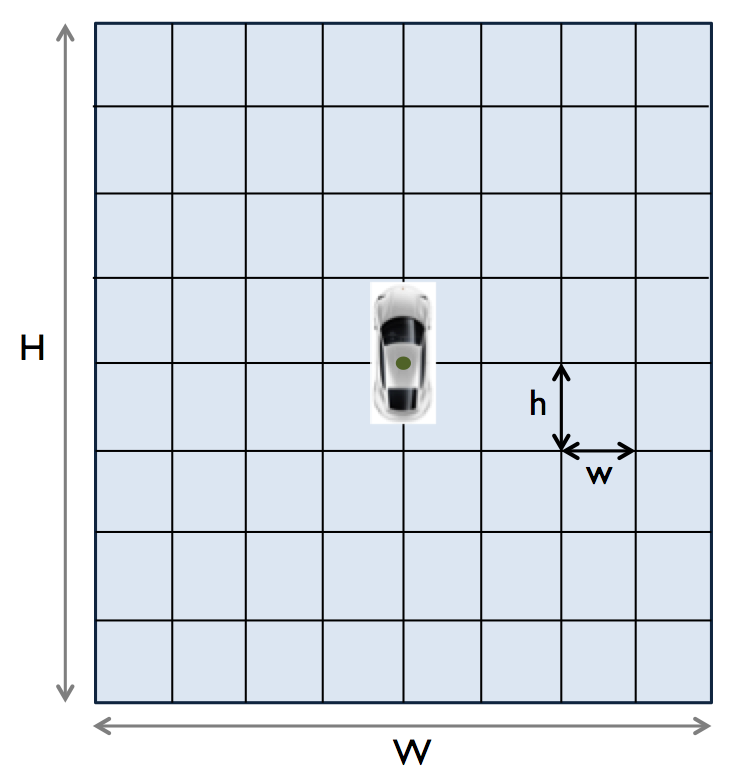}   
    \\
    {\small (a) 3D voxel-based representation.} &  {\small  (b) Range-view-based representation.} &  {\small  (c) Bird's-eye-view-based representation.}
  \end{tabular}
\end{center}
\caption{\label{fig:representations} Common approaches to discretize the 3D space. The 3D voxel-based representation is to discretize the 3D space into equally-spaced nonoverlapping voxels from each of the three dimensions; the range-view-based representation is to discretize the 3D space along the azimuth angle and the elevation angle; and the bird's-eye-view-based representation is to discretize
the 3D space  along the $X, Y$ axes, omitting the height dimension.}
\end{figure*}

\mypar{3D voxelization}
\label{sec:3dv}
To enjoy the success of 2D image processing and computer vision, we can discretize the 3D space into voxels and use a series of voxels to represent a 3D point cloud. A straightforward discretization is to partition the 3D space into equally-spaced nonoverlapping voxels from each of three dimensions; see Figure~\ref{fig:representations} (a).  Let a 3D space with range $H, W, D$ along the $X, Y, Z$ axes respectively. Each voxel is of size $h, w, d$, respectively.  The $(i, j, k)$th voxel represents a 3D voxel space, 
$
\V_{i,j, k}  \ = \ \{ (x, y, z)  |   (i-1) h  \leq x < i h, 
  (j-1) w  \leq y < j w, 
  (k - 1 ) d \leq z < k d \}.
$
We then use a three-mode tensor to represent this 3D point cloud. Let $\X^{\rm (vox)} \in \R^{H \times W \times D}$, whose  $(i, j, k)$th element is
\begin{equation}
\label{eq:voxel_based_representation}
 \X^{\rm (vox)}_{i,j,k}  \ = \
  \left\{ 
    \begin{array}{rl}
      1, & \mbox{when $\V_{i, j, k} \cap \S \neq \emptyset$};\\
      0, & \mbox{otherwise}.
  \end{array} \right. 
\end{equation}
The tensor $\X^{\rm (vox)}$ records the voxel occupancy.

The advantages of the 3D-voxelization-based representation are that (i) the resulting voxels are associated with a natural hierarchical structure and all the voxels have a uniform spatial size; and (ii) we can use off-shelf tools, such as 3D convolutions to analyze data; however, the shortcomings are that (i) it does not consider specific properties of organized 3D point clouds; (ii) it usually leads to an extremely sparse representation where most voxels are empty; and (iii) it involves a serious trade-off between the resolution and the memory. This representation can be used in the perception module of autonomous driving, as well as the storage of 3D point clouds.

\mypar{Range view}
\label{sec:range_view}
As discussed in Section~\ref{sec:properties}, a real-time LiDAR sweep is essentially a series of range measurements from a single location with certain angular field of view; see Figure~\ref{fig:representations} (b). We can approximately organize the 3D points in a real-time LiDAR to a 2D range-view image. Each pixel in the range-view image corresponds to a frustum in the 3D space. The pixel value is the range from the LiDAR to the closest 3D point inside the frustum.
Specifically, we partition the 3D space along the azimuth angle $\alpha \in [0 , 2 \pi )$ and the elevation angle $\theta \in (- \pi/2, \pi/2]$ with the resolution of azimuth angle $\alpha_0$ and the resolution of elevation angle $\theta_0$. The $(i, j)$th pixel corresponds to a frustum space, 
$
\V_{i,j}  \ = \ \{ (x, y, z)  |    \alpha_0 (i-1)  \leq  {\rm acos}(\frac{x}{\sqrt{x^2 + y^2}})  <  \alpha_0 i, 
  \theta_0 (j-1)  \leq   {\rm atan}(\frac{z}{\sqrt{x^2 + y^2}})  +  \frac{\pi}{2}    < \theta_0 j    \}.
$
We then use a 2D matrix to represent a 3D point cloud. Let $\X^{\rm (FV)} \in \R^{H \times W }$, whose $(i,j)$th element is
\begin{equation}
\label{eq:range_view_based_representation}
 \X^{\rm (FV)}_{i,j}  \ = \
  \left\{ 
    \begin{array}{rl}
      \min_{(x,y,z) \in \V_{i,j} \cap \S } \sqrt{x^2 + y^2 + z^2}, & \mbox{$\V_{i, j, k} \cap \S \neq \emptyset$};\\
      -1, & \mbox{otherwise}.
  \end{array} \right. 
\end{equation}
We consider the smallest range value in each frustum space. When no point falls into the frustum space, we set a default value as $-1$. Note that the range-view-based representation could also use nonuniform-spaced elevation angles according to the LiDAR setting.

The advantages of the range-view-based representation are that (i) it naturally models how LiDAR captures 3D points, reflecting a 2D surface in the 3D space; (ii)  Most frustum spaces associated have one or multiple 3D points, leading to a compact range-view image; however, the shortcoming is that it is difficult to model an unorganized point cloud, such as the point-cloud map in an HD map. This representation can be used in the perception module.

\mypar{Bird's-eye view}
\label{sec:bird_eye_view}
The bird's-eye-view (BEV)-based representation is a special case of 3D voxelization by ignoring the height dimension. It projects 3D voxels to a BEV image; see Figure~\ref{fig:representations} (c). Let a 3D space with range $H, W$ along the $X, Y$ axes respectively. Each pixel is of size $h, w$ respectively.  The $(i, j)$th pixel in the BEV image represents a pillar space, 
$
\V_{i, j}  \ = \ \{ (x, y, z)  |   (i-1) h  \leq x < i h, (j-1) w \leq y < j w \}.
$
We then use a 2D matrix to represent a 3D point cloud. Let $\X^{\rm (BEV)} \in \R^{H \times W }$, whose $(i, j)$th element is
\begin{equation}
\label{eq:bev_based_representation}
 \X^{\rm (BEV)}_{i,j}  \ = \
  \left\{ 
    \begin{array}{rl}
      1, & \mbox{when $\V_{i, j} \cap \S \neq \emptyset$};\\
      0, & \mbox{otherwise}.
  \end{array} \right. 
\end{equation}
The matrix $\X^{\rm (BEV)}$ records the occupancy in the 2D space. Note that there are a few variations of the BEV-based representations. For example, instead of using a binary value, MV3D~\cite{ChenMWLX:17} uses a few statistical values in each pillar space to construct $\X^{\rm (BEV)}$.

The advantages of the BEV-based representation are that (i) it is easy to apply 2D vision-based techniques; (ii) it is easy to merge with information from the HD map. For example, drivable areas and the positions of intersections encoded in the HD map can be projected to the same 2D space and fuse with LiDAR information;  (iii) it is easy to use for subsequent modules, such as prediction and motion planning, and (iii) objects are always the same size regardless of range (contrasting with the range-view-based representation), which is a strong prior and makes the learning problem much easier; however, the shortcoming of this voxelization is that 
(i) it also involves a serious trade-off between resolution and memory, causing excessive quantization issues of getting detailed information on small objects;
(ii) it does not consider the specific properties of organized 3D point clouds and cannot reason the occlusion; and (iii) it causes the sparsity issue because most pixels are empty. This representation can be used in the perception module of autonomous driving. 

\subsection{Representative tools}
\label{sec:tools}
3D point clouds have been studied across various communities, such as robotics, computer graphics, computer vision and signal processing. We introduce a few representative tools to process and learn from 3D point clouds. We mainly emphasize deep-neural-network-based approaches because of their practical usages in autonomous driving.

\mypar{Non-deep-learning methods} 
Before the emergence of deep learning, there have been many traditional methods to handle 3D point clouds for various tasks. However, unlike deep neural networks, those conventional methods can hardly be described in a single methodological framework. This is
because hand-crafted tools are specifically designed to cater to the needs of each individual task. For example, in 3D point cloud segmentation and 3D shape detection, traditional techniques have been developed based on either region growth with simple geometric heuristics, or graph-based optimization, or robust estimation methods, such as RANSAC~\cite{Fischler_RANSAC_CACM81}. As another important task, 3D keypoint matching is closely related to 3D point cloud registration and 3D point cloud recognition. To tackle this task, many statistics-based methods have been developed in a hand-crafted fashion and aim to describe the geometric structures around 3D keypoints or objects; see a more comprehensive discussion in~\cite{hana2018comprehensive}.

\mypar{Convolutional neural networks}
The motivation of using convolutional neural networks is to leverage off-shelf deep learning tools to process 3D point clouds. As regularized versions of multilayer perceptrons, convolutional neural networks (CNNs) employ a series of convolution layers and are commonly applied to analyzing images and videos. A convolution layer operates a set of learnable filters on input data to produce the output that expresses the activation map of filters.  The beauty of a convolution layer is weight-sharing; that is, the same filter coefficients (weights) are applied to arbitrary positions in a 2D image, which not only saves a lot of learnable weights, but also ensures shift invariance, and helps avoid overfitting to limited training data.  As a general and mature learning framework, CNNs and common variations are widely used in various computer vision tasks, including classification, detection, and segmentation, and have achieved state-of-the-art performance in most tasks. 

Based on the success of CNNs in images and videos, CNNs have been applied to 3D point cloud data as well. Multiple representations have been used, including the 3D-voxelization-based representation~\eqref{eq:voxel_based_representation}, the range-view-based representation~\eqref{eq:range_view_based_representation} and the BEV-based representation~\eqref{eq:bev_based_representation}. A benefit of using CNNs to handle a 3D point cloud is that a convolution operator naturally involves local spatial relationships. In PointNet, each 3D point is processed individually; while in CNNs, adjacent voxels or pixels are considered jointly, providing richer contextual information. The basic operator is a 3D convolution for the 3D voxelization-based representation and a 2D convolution for the range-view-based representation and the BEV-based representation, respectively. Without loss of generality, consider a 4-mode tensor $\X \in \R^{I \times J \times K \times C}$, after convolving with $C$ 3D filters $\Hh \in \R^{k \times k \times k \times C}$, the $(i,j,k,c')$th element of the output $\Yy \in \R^{I \times
J \times K  \times C'}$  is
$$
\Yy_{i,j,k,c'} \ = \ \sum_{c=0}^{C-1} \sum_{\ell=0}^{k-1} \sum_{m=0}^{k-1} \sum_{n=0}^{k-1}  \Hh_{i-\ell,j-m,k-n,c'}  \X_{\ell,m,n, c}.
$$
For simplicity, we omit the boundary issue.  3D convolution is expensive in   both computation and memory usage.

Because of the discretization, many techniques and architectures developed for 2D images can be easily extended to handle 3D point clouds. Even though the discretization causes inevitable loss of information, CNNs usually provide reliable performances and are widely used in many tasks. As discussed previously, one critical issue about discretizing a 3D point cloud is that a resulting 3D volume or 2D image is sparse. A huge amount of computation is wasted in handling empty voxels.

To summarize, CNNs handle a 3D point cloud in a discretized representation. This approach inevitably modifies the exact 3D position information, but still provides strong and promising empirical performances because of the spatial relationship prior and the maturity of CNNs. It is thus widely used in the industry.

\mypar{PointNet-based methods}
The motivation of using PointNet-based methods is to directly handle raw 3D points by deep neural networks without any discretization. PointNet~\cite{QiSMG:17} is a pioneering work that achieves this goal. Raw 3D point clouds are inherently unordered sets, and PointNet was designed to respect this property and produce the same output regardless of the ordering of the input data. The key technical contribution of PointNet is to use a set of shared point-wise multi-layer perceptrons (MLPs) followed by global pooling to extract geometric features while ensuring this permutation-invariant property of raw 3D data. Even though the architecture is simple, it has become a standard building block for numerous 3D point cloud learning algorithms and achieves surprisingly strong performance on 3D point cloud recognition and segmentation. 

\begin{figure*}[thb]
  \begin{center}
    \begin{tabular}{cc}
   \includegraphics[width=0.95\columnwidth]{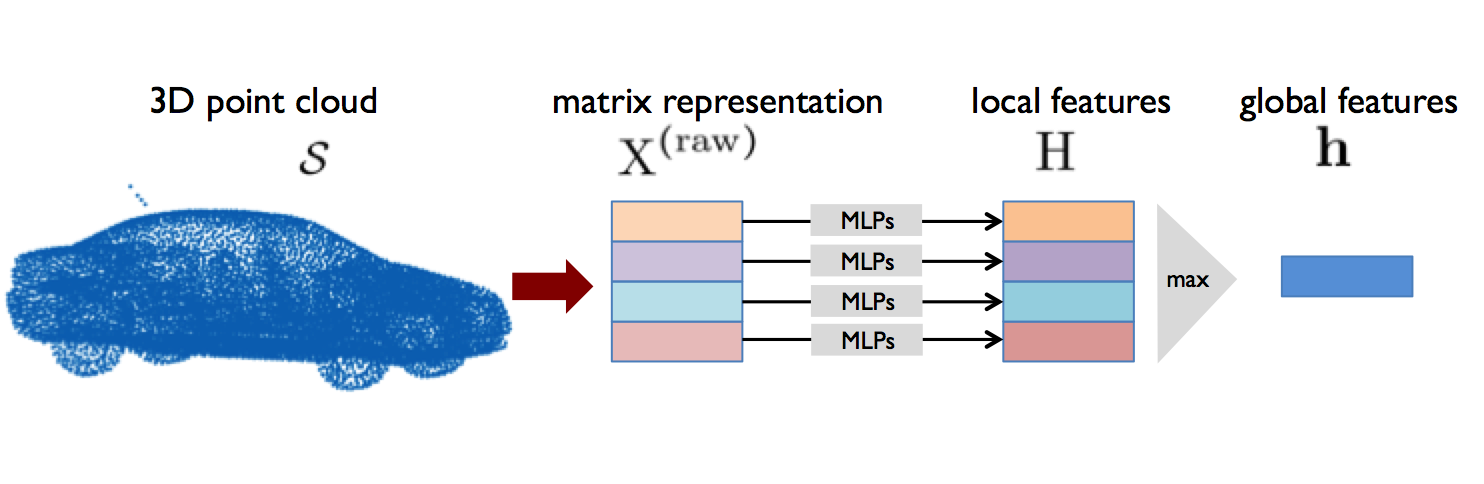}   
  & \includegraphics[width=0.8\columnwidth]{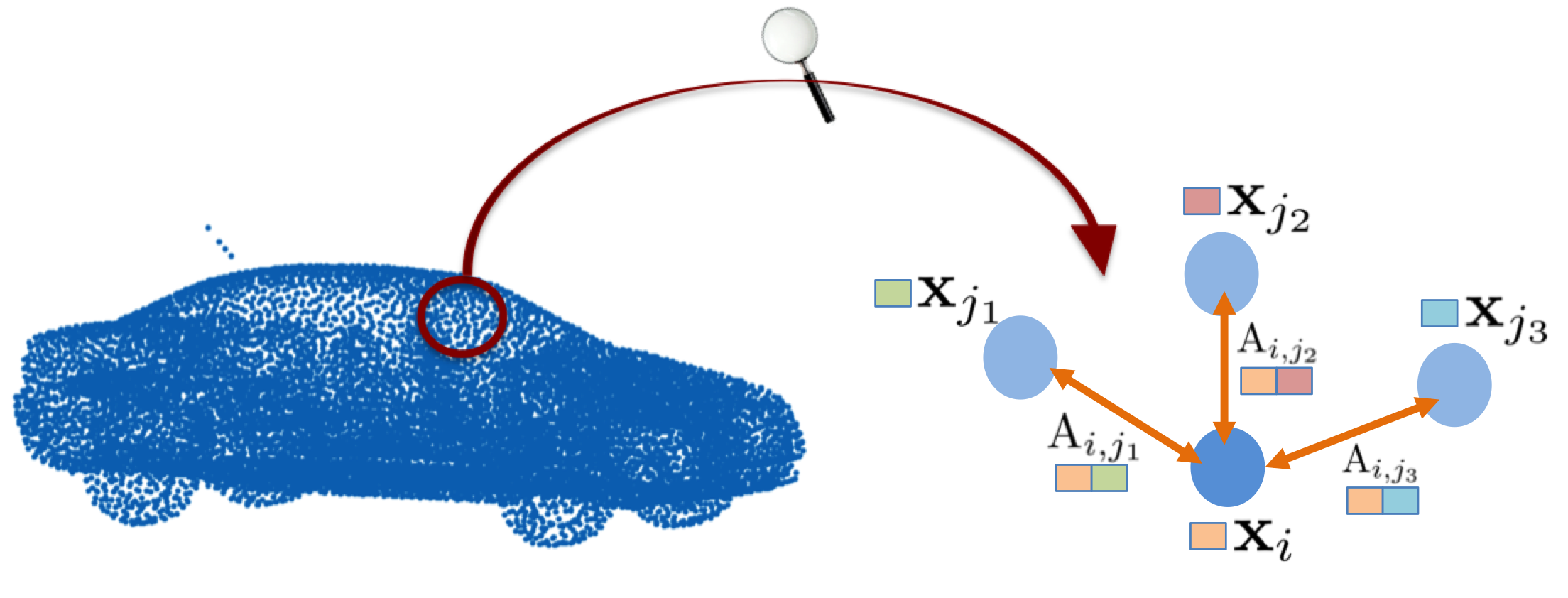}   
    \\
    {\small (a) PointNet.} &  {\small  (b) Graph-based methods.}
  \end{tabular}
\end{center}
\caption{\label{fig:tools} Illustration of representative tools. Plot (a) shows that PointNet uses a set of shared point-wise multi-layer perceptrons (MLPs) followed by max-pooling to extract geometric features that exhibit the permutation-invariant property of raw 3D point clouds. Plot (b) shows that graph-based methods introduce
  a graph structure to capture the local relationships among 3D points. In the graph, each node is a 3D point and each edge reflects the relationship between each pair of 3D points.}
\end{figure*}

PointNet considers the raw-point-based representation $\X^{(\rm raw)}$~\eqref{eq:raw_point_representation}. Let $\Hh \in \mathbb{R}^{N \times D}$ be a local-feature matrix, where the $i$th row $\Hh_i$ represents the features for $i$th point, and $\mathbf{h} \in \mathbb{R}^{D}$ be a global-feature vector. A basic computational block of PointNet works as
\begin{eqnarray}
\label{eq:PointNet}
\Hh_i  & = & {\rm MLP^{(L)}} \left( \X^{(\rm raw)}_i  \right) \in \mathbb{R}^{D}, ~~~{\rm for}~i = 1, \cdots, N, 
\\ \nonumber
\label{eq:encoder_2}
\mathbf{h} & = & {\rm maxpool} \left( \Hh \right) \in \mathbb{R}^{D},
\end{eqnarray}
where $\X^{(\rm raw)}_i$ is the $i$th 3D point's feature, and ${\rm MLP^{(L)}}(\cdot)$ denotes a $L$-layer MLPs, which map each 3D point to a feature space, and ${\rm maxpool}(\cdot)$ performs downsampling by computing the maximum values along the column (the point dimension); see Figure~\ref{fig:tools} (a). Note that each 3D point goes through the same MLPs separately.

Intuitively, the MLPs propose $D$ representative geometric patterns and test if those patterns appear around each 3D point. The max-pooling records the strongest response over all the 3D points for each pattern. Essentially, the global-feature vector $\mathbf{h}$  summarizes the activation level of  $D$ representative geometric patterns in a 3D point cloud, which can be used to recognize a 3D point cloud. Meanwhile, since each 3D point goes through the same MLPs separately and the max-pooling removes the point dimension, the entire computational block is permutation invariant; that is, the ordering of 3D points does not influence the output of this block.  To some extent, PointNet for 3D point cloud learning 
is similar to principal component analysis (PCA) for data analysis: it is simple, general and effective. Just like principal component analysis, PointNet extracts global features in a 3D point cloud. 

To summarize, PointNet-based methods handle 3D point clouds in the raw-point-based representation and ensure the permutation invariance. The effectiveness has been validated in various processing and learning tasks.

\mypar{Graph-based methods}
The motivation of using graph-based methods is to leverage the spatial relationships among 3D points to accelerate the end-to-end learning of deep neural networks. One advantage of CNNs is that a convolution operator considers local spatial relationships; however, those relationships are between adjacent voxels (or adjacent pixels), not original 3D points. To capture the local relationships among 3D points, one can introduce a graph structure, where each node is a 3D point and each edge reflects the relationship between each pair of 3D points. This graph structure is a discrete proxy of the surface of an original object. A matrix representation of a graph with $N$ nodes is an adjacency matrix $\Adj \in \R^{N \times N}$, whose $(i,j)$th element reflects the pairwise relationship between the $i$th and the $j$th 3D points; see Figure~\ref{fig:tools} (b). Graph-based methods usually consider the raw-point-based representation~\eqref{eq:raw_point_representation}. Each column vector in $\X^{\rm (raw)}$ is then data supported on the graph $\Adj$; called a graph signal.

There are several ways to construct a graph, such as a $K$-nearest-neighbor graph, an $\epsilon$-graph and a learnable graph. A $K$-nearest-neighbor graph is a graph in which two nodes are connected by an edge, when their Euclidean distance is among the $K$-th smallest Euclidean distances from one 3D point to all the other 3D points. An $\epsilon$-nearest-neighbor graph is a graph in which two nodes are connected by an edge, when their Euclidean distance is smaller than a given threshold $\epsilon$. Both $K$-nearest-neighbor graphs and $\epsilon$-graphs can be efficiently implemented by using efficient data structures, such as Octree~\cite{PengK:05}. A learnable graph is a graph whose adjacency matrix is trainable in an end-to-end learning architecture.

A general graph-based operation  is a graph filter, which extends a classical filter to the graph domain and extracts features from graph signals. The most elementary nontrivial graph filter is called a~\emph{graph shift operator}. Some common options for a
graph shift operator include the adjacency matrix $\Adj$,
the transition matrix $\D^{-1} \Adj$ ($\D$ is the weighted degree
matrix, a diagonal matrix with $\D_{i,i} = \sum_{j} \Adj_{i,j}$
reflecting the density around the $i$th point), the graph Laplacian
matrix $\D-\Adj$, and many other structure-related matrices; see details in~\cite{OrtegaFKMV:18}. The graph
shift replaces the signal value at a node with a weighted linear
combination of values at its neighbors; that is, $ \Yy \ = \ \Adj \X^{(\rm raw)} \
\in \ \R^N, $ where $\X^{(\rm raw)} \in \R^{N \times 3}$ is an input graph signal (an
attribute of a point cloud). Every linear, shift-invariant graph filter is a polynomial in the graph shift,
$$
  h(\Adj) \ = \ \sum_{\ell = 0} ^{L-1}  h_{\ell} \Adj^{\ell} = h_0\Id + h_1\Adj + \ldots + h_{L-1}\Adj^{L-1},
$$
where $h_{\ell}$, $\ell = 0, 1, \ldots, L-1$ are filter coefficients
and $L$ is the graph filter length. A higher order corresponds to a larger receptive field on the graph vertex domain. The output of graph filtering is given by the matrix-vector product $ \Yy = h(\Adj) \X^{(\rm raw)}$. Graph filtering can be used in various processing tasks, such as 3D point cloud downsampling and denoising~\cite{ChenTFVK:18}.

Inspired by the success of graph neural networks in social network analysis, numerous recent research incorporate  graph neural networks to handle a 3D point cloud. As the first such work,~\cite{WangSLSBS:18} introduces two useful techniques: the edge convolution operation and learnable graphs. The edge convolution is a convolution-like operation to extract geometric features on a graph. The edge convolution exploits local neighborhood information and can be stacked to learn global geometric properties.  Let $\Hh \in \mathbb{R}^{N \times d}$ be a local-feature matrix, where the $i$th row $\Hh_i$ represents the features for the $i$th point. A basic computational block works as
$
\Hh_i = \|_{(i,j) \in \E} g (\X^{(\rm raw)}_i, \X^{(\rm raw)}_j) \in \R^d,
$
where $\E$ is the edge set and $g(\cdot, \cdot)$ is a generic mapping, implemented by some neural networks, and $\|$ is a generic aggregation function, which could be the summation or maximum operation. To some extent, the edge convolution extends PointNet by inputting a pair of neighboring points' features. The edge convolution is also similar to graph filtering: both aggregates neighboring information; however, the edge convolution  specifically models each pairwise relationships by a nonparametric function.~\cite{WangSLSBS:18} also suggests to dynamically learn a graph. It always uses a kNN graph, but the distance metric is the Euclidean distance in the high-dimensional feature space. The edge convolution can be reformulated as the continuous convolution in the 3D space, which ensures shift-invariance~\cite{ChenNLL:19}.

Subsequent research has proposed to use novel graph neural networks to handle 3D point cloud recognition and segmentation. As one of the most recent works in this area,~\cite{LiMTG:19} constructs the deepest yet graph convolution network (GCN) architecture, which has 56 layers. It transplants a series of techniques from CNNs, such as residual and dense connections, and dilated graph convolutions, to the graph domain.

To summarize, graph-based methods build graph structures to capture the distribution of a 3D point cloud and take advantage of local spatial relationships. This approach handles 3D point clouds in the raw-point-based representation, ensuring the permutation invariance. This approach is less mature: even though leveraging a graph improves the overall performance, graph construction is more art than science and takes extra computational cost~\cite{WangSLSBS:18}; additionally, deep architectures for graph-based neural networks still needs more exploration~\cite{LiMTG:19}.

\section{3D Point Cloud Processing for High-Definition Map Creation} 
\label{sec:map}
\subsection{Overview of high-definition map creation module}
A high-definition (HD) map for autonomous driving is a precise heterogeneous map representation of the static 3D environment and traffic rules. It usually contains two map layers: a point-cloud map, representing 3D geometric information of surroundings, and a traffic-rule-related semantic feature map, containing road boundaries, traffic lanes, traffic signs, traffic lights, the height of the curbs, etc. The main reason for creating an offline HD map is that understanding traffic rules in real-time is too challenging. For example, based on the current technology, it is difficult for an autonomous vehicle to determine the correct lane in real-time when driving into at an intersection with complicated lane merging and splitting. In contrast, all traffic rules and environmental information can easily be encoded in an HD map, which goes through an offline process with human supervision and quality assurance. An HD map provides strong and indispensable priors and fundamentally eases the designs of multiple modules in an autonomy system, including localization, perception, prediction and motion planning. Therefore, an HD map is widely believed to be an indispensable component of autonomous driving.

\mypar{Priors for localization}
The role of localization is to localize the
pose of an autonomous vehicle. In an HD map, the point-cloud map and the traffic-rule-related semantic features, such as lane markers and poles, are usually served as localization priors for the map-based localization. These priors are used to register real-time LiDAR sweeps to the point-cloud map, such that one can obtain the real-time high-precision ego-motion of an autonomous vehicle. 

\mypar{Priors for perception}
The role of perception is to detect all objects in the scene, as well as their internal states. The perception module can use an HD-map to serve as a prior for detection. For example, the positions of traffic lights in an HD map are usually served as perception priors for traffic light state estimation. With the point-cloud map as priors, one can separate a real-time LiDAR sweep into foreground and background points in real-time. We can then remove background points, which are those lying on the static scenes, such as road surfaces and the trunks of trees, and feed only foreground points to the perception module. This formalism can significantly reduce the computational cost and improve the precision of object detection.

\mypar{Priors for prediction}
The role of prediction is to predict the future trajectory of each object in the scene.  In an HD map, 3D road and lane geometries and connectivities are important priors to the prediction module. These priors can be used to guide the predicted trajectories of objects to follow the traffic lanes.

\mypar{Priors for motion planning} 
The role of motion planning is to determine the trajectory of an autonomous vehicle. In an HD map, traffic-rule-related semantic features such as lane geometries and connectivities, traffic light, traffic sign and the speed limit of lanes, are indispensable priors for the motion-planning module.  These priors are used to guide the designed trajectory to follow the correct lane and obey the stop signs and other traffic signs.

\begin{figure*}[thb]
  \begin{center}
   \includegraphics[width=1.8\columnwidth]{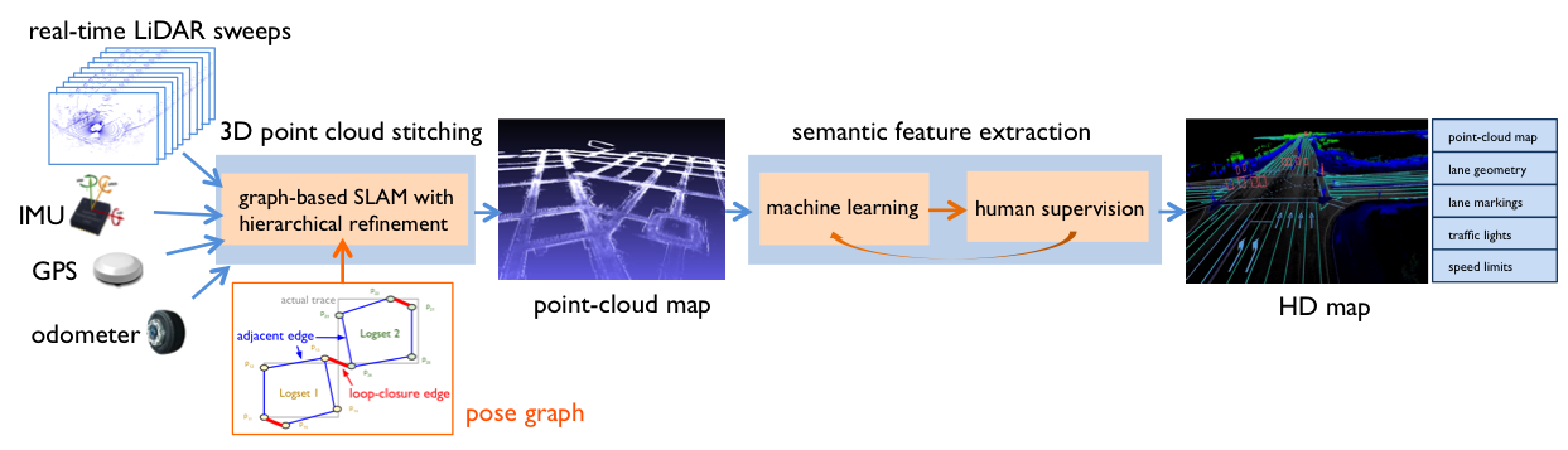}
\end{center}
\caption{\label{fig:hdmap_creation} A standard HD map creation system includes two core components: 3D point cloud stitching and semantic feature extraction. 3D point cloud stitching usually adopts graph-based SLAM with hierarchical refinement; and semantic feature extraction contains iterative procedures of machine learning and human supervision. A key component in graph-based SLAM is a pose graph, modeling the relations among LiDAR poses. The nodes are LiDAR poses and edges reflecting the misalignment level between two LiDAR poses. The final outputs include a point-cloud map, which is a dense 3D point cloud, as well as a traffic-rule-related semantic feature map, containing the positions of landmarkers, traffic signs and traffic lights.
}
\end{figure*}

Since an HD map is critical to autonomous driving, it must be created with high precision and be up-to-date. To achieve this, it usually needs sophisticated engineering procedures to analyze data from multiple modalities by exploiting both machine learning techniques and human supervision. A standard map creation module includes two core components:~\emph{3D point cloud stitching} and~\emph{semantic feature extraction}; see Figure~\ref{fig:hdmap_creation}. 3D point cloud stitching merges real-time LiDAR sweeps collected from multiple vehicles across times into a point-cloud map; and semantic feature extraction extracts semantic features, such as lane geometries and traffic lights, from the point-cloud map. See a video illustration of the industrial-level HD maps through the link\footnote{\url{https://vimeo.com/303412092}}  and additional illustrations in Appendix.

\subsection{3D point cloud stitching}
The goal of 3D point cloud stitching is to create a high-precision point-cloud map from the sensor data collected by a fleet of vehicles across time. Since a point-cloud map dominates the precision of all the map priors, centimeter-level precision is required for any local portion of the point-cloud map. To promptly create and update city-scale HD maps, the process of 3D point cloud stitching  must be highly robust and efficient.  

One fundamental problem of 3D point cloud stitching is to estimate the 6-degree-of-freedom (DOF) pose of each LiDAR sweep; also called LiDAR pose. We consider the map frame as the standardized global frame, and the LiDAR frame as the ego frame of an autonomous vehicle at the time stamp when the corresponding real-time LiDAR sweep is collected. A LiDAR pose is then a transformation between the map frame and the LiDAR frame. It includes 3D translation and 3D rotation. Note that the 6-DOF pose can be represented as a 4$\times$4 homogeneous transformation matrix. With the LiDAR poses, all the LiDAR sweeps can be synchronized to the standardized global frame and integrated to form a dense 3D point cloud. To estimate LiDAR poses, a common technique is simultaneous localization and mapping (SLAM).
Let $\S_i$ and $\S_j$ be the $i$th and $j$th real-time LiDAR sweeps, respectively. SLAM works as
\begin{equation}
\label{eq:slam}
\text{argmin}_{p}\left[\sum_{p_i}\sum_{p_j} h_{\S_i, \S_j}(p_i,p_j) + g(p_i)\right],
\end{equation}
where $p_i$ is the 6-DOF LiDAR pose associated to the $i$th real-time LiDAR sweep, $h_{\S_i, \S_j}(p_i,p_j)$ indicates the negative log likelihood of the measurement on the misalignment between $\S_i$ and $\S_j$, and $g(\cdot)$ indicates the negative log likelihood of the difference between the predicted LiDAR position in the map frame and the direct measurement of GPS~\cite{GrisettiKSB:10}. A typical choice of $h_{\S_i, \S_j}(p_i,p_j)$ is the objective function of the iterative closest point (ICP) algorithm. We thus minimize the objective function of the ICP algorithm and assign the optimized value to $h_{S_i, S_j}(p_i,p_j)$.

SLAM is a big research field in robotics communities and there exists extensive research that aims to solve the optimization problem~\eqref{eq:slam}. For example, the filter-based SLAM solves the optimization problem~\eqref{eq:slam} in an approximated and online fashion. It employs Bayes filtering to predict and optimize the map and LiDAR poses iteratively based on the online sensor measurements. On the other hand, the graph-based SLAM optimizes all the LiDAR poses together by using all sensor measurements across time. It constructs a pose graph that models the relations among LiDAR poses, In the pose graph, the $i$th node is the $i$th LiDAR pose, $p_i$; and the $(i,j)$th edge is the cost of misalignment between the $i$th and $j$th LiDAR poses, $h_{S_i, S_j}(p_i,p_j)$; see the pose graph in Figure~\ref{fig:hdmap_creation}. Intuitively, each edge weight is either the total point-to-point distance or the total point-to-plane distance between two LiDAR sweeps. Solving~\eqref{eq:slam} is thus equivalent to minimizing the total sum of the edge weights of a pose graph. 

For a city-scale map creation, the SLAM solution must satisfy the following requirements.

\mypar{High local and global precision}~\emph{Local precision} indicates that the LiDAR poses in a local region are accurate with respect to one another; and ~\emph{global precision} indicates that all the LiDAR poses in the entire HD map are accurate with respect to the standardized global frame. For the SLAM solution, centimeter/micro-radian level local precision must be achieved because autonomy software modules require the highly accurate local surroundings from the HD map; and the centimeter-level global precision is useful to accelerate the HD map update process especially for the city-scale application;

\mypar{High robustness} The SLAM solution requires to handle the noisy sensor measurements collected by multiple vehicles driving in complicated scenes and complex driving conditions in the real world; and
    
\mypar{High efficiency} The SLAM solution requires to handle the optimization of over $100$ millions of LiDAR poses.

To achieve high precision and robustness, the graph-based SLAM is a better option than the filter-based SLAM because the global optimization formalism makes the graph-based SLAM inherently more accurate; however, it is still challenging to solve the city-scale graph-based SLAM problem with high efficiency and robustness. There are two main reasons. First, the scale of the problem is enormous. It is expensive to solve the optimization problem~\eqref{eq:slam} in a brute-force way because the core step of the optimization algorithm is to solve a series of equation associated with an $n$-by-$n$ matrix, where $n$ is the total number of LiDAR poses. For a city-scale map, $n$ could be more than $100$ millions, causing big issues for both computational efficiency and numerical stability of the optimization algorithm. Second, evaluating edge weights in a pose graph usually suffers from low precision because sensor data is collected in complex driving conditions. For example, the calculation of the misalignment between consecutive LiDAR sweeps will likely be compromised by the moving objects.

To effectively solve this problem, the graph-based SLAM with the hierarchical refinement formalism can be adopted~\cite{DroeschelB:18}.  The functionality of hierarchical refinement formalism is to provide  a good initialization for the global optimization, making the optimization both fast and accurate. The hierarchical refinement formalism distinguishes two types of edges in a pose graph; that is, adjacent edges and loop-closure edges. Adjacent edges model the relations between two LiDAR poses whose corresponding LiDAR sweeps are consecutively collected from the same logset; and loop-closure edges model the relations between two LiDAR poses whose corresponding LiDAR sweeps are collected around the same location from different logsets (different vehicles or across time). To handle these two types of edges,  the hierarchical refinement formalism includes two steps: (1) optimizing adjacent edges, including a chain of LiDAR poses from a single logset; and (2) optimizing loop-closure edges, including LiDAR poses across logsets; see Figure~\ref{fig:hdmap_creation}. In the first step, rather than relying simply on aligning LiDAR sweeps, sensor  measurements from multiple modalities, including IMU, GPS, odometer, camera and LiDAR, can be fused together to calculate the adjacent edges.
Because consecutive LiDAR sweeps have similar LiDAR poses, this step is usually easy and provides extremely high precision. In the second step, the loop-closure edges are calculated by aligning LiDAR sweeps through the ICP algorithm. After these two steps, we then perform the global optimization~\eqref{eq:slam}. 

Since most edges in a pose graph are adjacent edges, which can be highly optimized through the first step, the hierarchical refinement formalism provides a good initialization for the global optimization. Therefore,  the computational cost for optimizing the entire pose graph can  be significantly reduced and the robustness of the global optimization can be greatly improved by the hierarchical refinement formalism.

\subsection{Semantic feature extraction}
The goal of semantic feature extraction is to extract traffic-rule-related semantic features, such as lane geometries, lane connectivities, traffic signs and traffic lights, from the point-cloud map. This component requires both high precision and recall. For example, missing a single traffic light prior in a city-scale HD map can potentially cause serious issues to the perception and motion planning modules, which can severely jeopardize the safety of autonomous driving.

The semantic feature extraction component usually contains two iterative steps. The first step uses machine learning techniques to automatically extract features; and the second step introduces human supervision and quality assurance process to ensure the high precision and recall of the semantic features.

To automatically extract features, standard machine learning techniques are based on convolutional neural networks. The inputs are usually the combination of the LiDAR ground images and the camera images associated with the corresponding real-time LiDAR sweep. A LiDAR ground image renders the BEV-based representation of the point-cloud map obtained in 3D point cloud stitching, where the values of each pixel are the ground height and laser reflectivity of each LiDAR point. The outputs are usually the semantic segmentation of either the LiDAR ground images or the camera images. The networks follow from standard image segmentation architectures.

After obtaining the output, the pixel-wise semantic labels are projected back to the point-cloud map. By fitting the projected 3D points into 3D splines or 3D polygons, the traffic-rule-related semantic feature map can then be obtained. Note that the human-editing outcomes also serve as an important source of training data for automatic feature extraction algorithms, where these two steps therefore form a positive feedback loop to improve the precision and efficiency of HD map production.

\subsection{Real-world challenges}
There still exist several challenges for the HD map creation.

\mypar{Point-cloud map with centimeter-level global precision} Global precision can greatly benefit the updating of a city-scale point-cloud map.  The changes of the urban appearance usually take place locally. Ideally the map update should focus on the targeted portion of the pose graph; however, a point-cloud map with high local precision but without high global precision cannot freely access the targeted portion from a global aspect and  guarantee its precision. In comparison, given a point-cloud map with high global precision, one can focus on updating the targeted portion of the pose graph, thus significantly reducing the scale of computation; however, it is challenging to enforce the global precision to the graph-based SLAM. This is because the global optimization formalism of graph-based SLAM tends to distribute the error of each edge uniformly in the graph. Therefore, even if the GPS observations are accurate, the corresponding LiDAR poses can be misaligned after global optimization. Enforcing centimeter-level global precision of a point-cloud map can be especially challenging in the places where the GPS signal is  unavailable, such as in building canyon, tunnel and underground garage.
      
\mypar{Automatic semantic feature extraction} Although there exists extensive research  on the semantic segmentation based on 3D point clouds and camera images, it is still challenging to automatically extract the lane connectivities in intersections and traffic lights that indicate lane control relations. This is due to limited training labels and complex traffic conditions. Currently, the solution to extracting the complex semantic features such as traffic light to lane control information still relies largely on human supervision, which is both expensive and time-consuming.

\section{3D Point Cloud Processing for Localization}
\label{sec:localization}

\subsection{Overview of localization module}

\begin{figure*}[thb]
  \begin{center}
   \includegraphics[width=1.8\columnwidth]{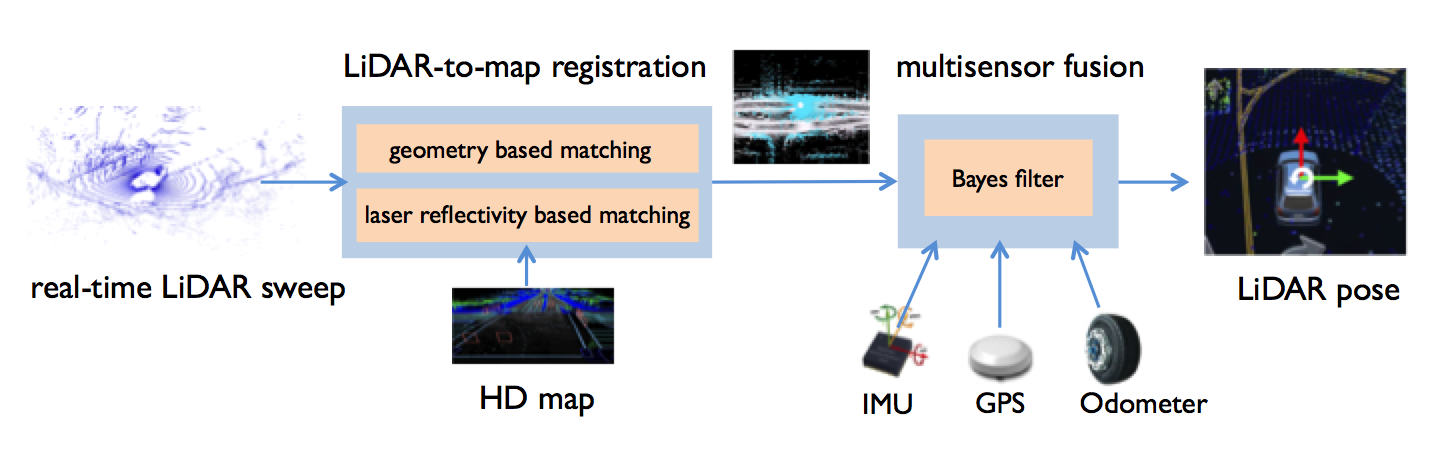}     
\end{center}
\caption{\label{fig:localization} A standard map-based localization system includes two core components: LiDAR-to-map registration and multisensor fusion. LiDAR-to-map registration uses geometry based matching and laser reflectivity based matching to achieve high precision and recall; and multisensor fusion adopts Bayes filters to
merge multiple modalities.}
\end{figure*}

As introduced in Section~\ref{sec:intro_autonomous_system}, the localization module finds ego position of an autonomous vehicle relative to the reference position in the HD map. It consumes the real-time measurements from multiple sensors, including LiDAR, IMU, GPS, odometer, cameras, as well as the HD map; see Figure~\ref{fig:localization}. Because of the 3D representation of an HD map, the ego position of an autonomous vehicle is a 6DOF pose (translation and rotation), which is a rigid transformation between the map frame and the LiDAR frame. The importance of the localization module to autonomous driving is that it bridges the HD map to the other modules in an autonomy system. For example, by projecting the HD map priors, such as the lane geometries to the LiDAR frame, the autonomous vehicle gains the knowledge of which lane itself drives on and which lanes the detected traffics are on. See a video illustration of the real-time localization through the link\footnote{\url{https://vimeo.com/327949958}} and additional illustrations in Appendix.

To enable the full autonomous driving, high precision and robustness are the most critical criteria for the performance of localization module. High precision indicates the error of translation should be at the centimeter level and the error of rotation angle should be at the micro-radian level. It allows the traffic detected from 1 kilometer away to be associated to the correct lanes in HD map, and the lane-change intentions of the closer traffic can be predicted by measuring the distance between its wheels to the lane boundaries, which can significantly benefit motion planning and prediction modules; and robustness indicates that the localization module is expected to work in all driving conditions with the changes of illumination, weather, traffic and the condition of roads. Note that although the commercial-grade GPS/IMU unit with real-time kinematics mode has accurate position measurement in open areas, it is not robust enough for autonomous driving because it suffers from the low precision issue in the city due to the multi-path effects.

To achieve these aforementioned criteria, the map-based localization with multi-sensor fusion is the standard approach. As discussed in previous sections, an HD map could be created beforehand and significantly ease the localization. On the contrary, the SLAM-based solution cannot satisfy these criteria.

\subsection{Map-based localization}
The basic idea of the map-based localization is to estimate the LIDAR pose by matching a LiDAR sweep to the point-cloud map in an HD map by leveraging the measurements from IMU, GPS, cameras to make pose estimation robust. A map-based localization system usually consists of two components; see Figure~\ref{fig:localization}. The first component is the LiDAR-to-map registration, which computes the LiDAR pose by registering LiDAR sweep to a point-cloud map; The second component is the multisensor fusion, which estimates the final pose from IMU, odometer, GPS, as well as the estimation from the LiDAR-to-map registration.

\mypar{LiDAR-to-map registration}
The LiDAR-to-map registration component is to directly estimate the LiDAR pose by matching the LiDAR sweep to the the point-cloud map. Let $\S$, ${\S^{(\rm map)}}$ be a real-time LiDAR sweep and the point-cloud map, respectively. The problem of LiDAR-to-map registration can be formulated as
\begin{equation}
\label{eq:icp}
\text{argmin}_p \left[ \sum_{\x_i \in \S} g \bigg(f_p(\x_i), {\S^{(\rm map)}}_{i^*} \bigg) \right],
\end{equation}
where $p$ is the LiDAR pose, $\x_i$ is the $i$th 3D point in the LiDAR sweep and ${\S^{(\rm map)}}_{i^*}$ is the 3D point in the point-cloud map that is associated with the $i$th 3D point in the LiDAR sweep. The associated index $i^*$ is usually chosen from the closest point in the Euclidean distance. The function $f_p: R^3 \rightarrow R^3$ is the function that transforms a 3D point $\x_i$ in the LiDAR frame into the map frame based on the LiDAR pose $p$; and the function $g(\cdot)$ indicates a loss function measuring the misalignment between the points from the LiDAR sweep and the HD map. Usually, $g(\cdot)$ takes the forms of the point-to-point, point-to-line, or point-to-plane distance between the associated points in the LiDAR sweep and the point-cloud map. 

To solve~\eqref{eq:icp} and achieve high precision and recall, there exist two major approaches.
\begin{itemize}
    \item \emph{Geometry based matching.}
    This approach calculates the high precision 6DOF pose by matching the LiDAR sweep to the point-cloud map based on the ICP algorithm~\cite{BeslM:92}. This approach usually works well in heavy traffic and challenging weather conditions, such as snow, because a point-cloud map contains abundant geometry priors for LiDAR sweeps to match with; however, in geometry-degenerated scenes, such as tunnel, bridge, and highway, the ICP calculation could diverge because of the loss of geometric patterns, hence causing bad precision; and
    
    \item \emph{Laser reflectivity based matching.}
    This approach calculates the pose by matching a LiDAR sweep to a point-cloud map based on laser reflectivity signals. The matching can be done in either the dense 2D image matching method or the feature extraction based ICP matching method. For the first method, the laser reflectivity readings of the LiDAR sweep and the point-cloud map are first converted into grey-scale 2D images, following the BEV-based representation~\eqref{eq:bev_based_representation}, and then the pose is calculated by image matching techniques. Note that this method only calculates the x, y, yaw components of the pose. To obtain the 6-DOF pose, the z, roll, pitch components are estimated based on the terrain information in the HD map. For the second method, the region of interest objects, such as lane markers,
    and poles, are firstly extracted from the LiDAR sweep based on the Laser reflectivity readings~\cite{HataW:14}. The ICP algorithm can then be used to calculate the LiDAR pose by matching the region of interest objects between a real-time LiDAR sweeps and the priors in the HD map. This approach usually outperforms the geometry based matching in the scenarios of highway and bridge, because those scenarios lack  geometry features but have rich laser reflectivity textures on the ground (e.g. dashed lane markers). This approach does not work well in the challenging weather conditions such as  heavy rain and snow where the laser reflectivity of the ground will change significantly.
\end{itemize}

To achieve the best performance, both of these two strategies can simultaneously be used to estimate LiDAR poses; however, LiDAR-to-map registration alone cannot guarantee the 100\% precision and recall for the pose estimation over the time. To give an extreme example, if LiDAR is totally occluded by trucks driving side-by-side or front-and-back, the LiDAR-to-map registration component would fail. To handle extreme cases and make the localization module robust, the multisensor fusion component is required. 

\mypar{Multisensor Fusion}
The multisensor fusion component is to estimate a robust and confident pose from measurements of multiple sensors, including IMU, GPS, odometer, cameras, as well as the poses estimated by the LiDAR-to-map registration module. The standard approach of multisensor fusion is to employ a Bayes-filter formalism, such as Kalman filter, extended Kalman filter, or particle filter. Bayes filters consider an iterative approach to predict and correct the LiDAR pose and other states based on the vehicle motion dynamics and the multisensor readings. In autonomous driving, the states tracked and estimated by Bayes filters usually include motion related states such as pose, velocity, acceleration, etc., and sensor related states such as IMU bias etc. 

Bayes filters work in two iterative steps: prediction and correction. In the prediction step, during the gaps between sensor readings, the Bayes filter predicts the states based on the vehicle motion dynamics and the assumed sensor model. For example, by taking the constant acceleration approximation as the vehicle motion dynamics during a short period of time, the evolution of pose, velocity, and acceleration can be predicted by Newton's laws. The IMU bias states can be predicted by assuming that it behaves as white noise. 

In the correction step, when receiving a sensor reading or a pose measurement, the Bayes filter corrects the states based on the corresponding observation models. For examples, when an IMU reading is received, the states of acceleration, angular velocities, and the IMU bias are corrected. When a pose measurement is received, the pose state is corrected. Note that the states require the correction because the prediction step is not prefect and there are accumulated errors over time. 

\subsection{Real-world challenges}
The real-world challenges of the localization module is to work in extreme scenes. For example, when an autonomous vehicle drives through a straight tunnel without dashed lane marker, there are few geometric and texture features, causing the failure of the LiDAR-to-map registration; when an autonomous vehicle is surrounded by large trucks, LiDAR could be totally blocked, also causing the failure of the LiDAR-to-map registration. When the failure of the LiDAR-to-map registration lasts for several minutes, the LiDAR pose estimated by the multisenor fusion component will drift significantly and the localization module will lose the precision.

\section{3D Point Cloud Processing for Perception}
\label{sec:perception}

\subsection{Overview of perception module} 
As introduced in Section~\ref{sec:intro_autonomous_system},
the perception module is the visual system of an autonomous vehicle that enables the perception of the surrounding 3D environment. The input of the perception module usually includes the measurements from cameras, LiDAR, RADAR and ultrasound, as well as the ego-motion pose output from the localization module and the priors from the HD map. The outputs of the perception module are typically traffic light states and objects' 3D bounding boxes with tracks. 

\begin{figure*}[thb]
  \begin{center}
    \begin{tabular}{ccc}
   \includegraphics[width=0.95\columnwidth]{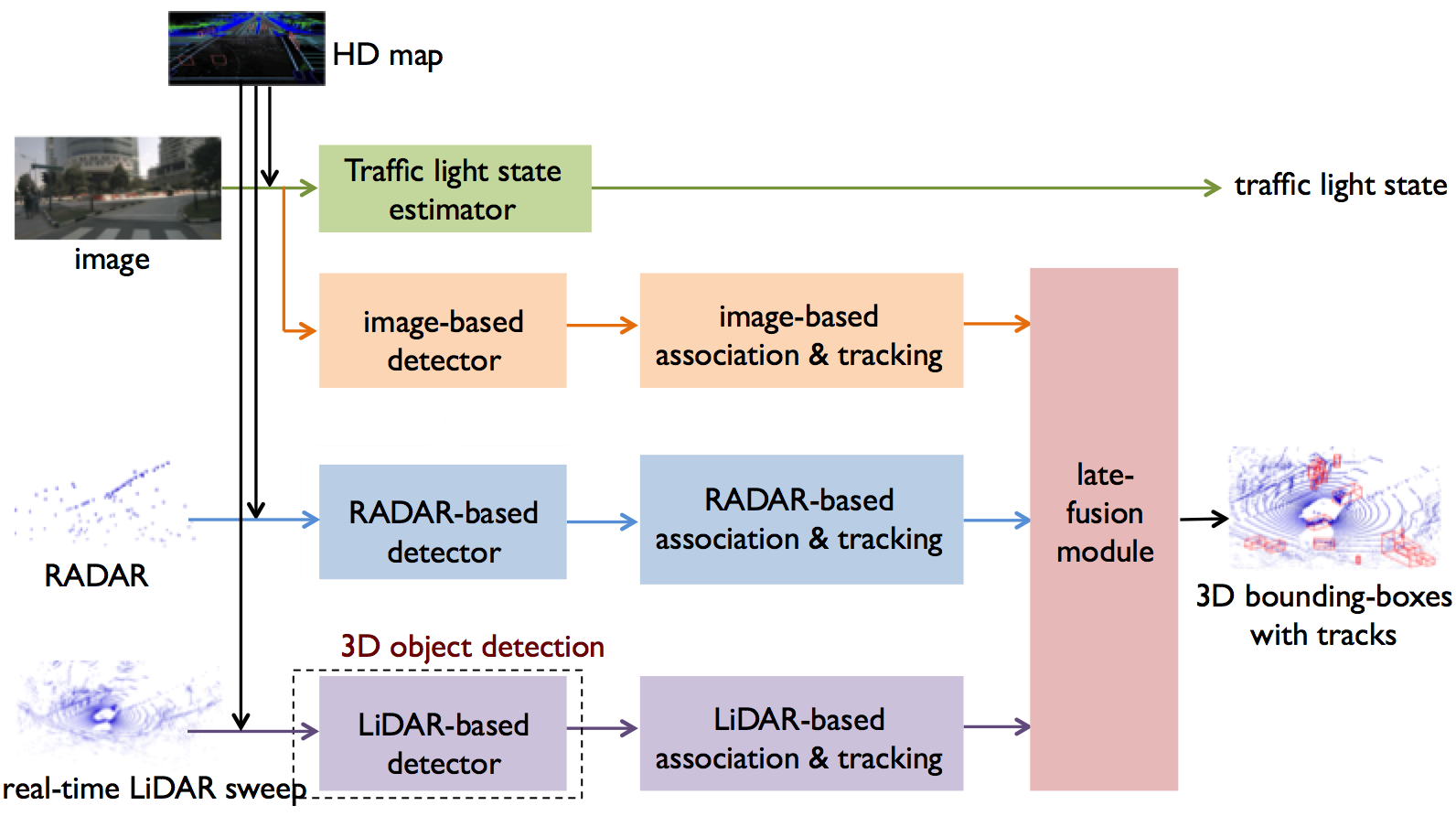}   &  & \includegraphics[width=0.82\columnwidth]{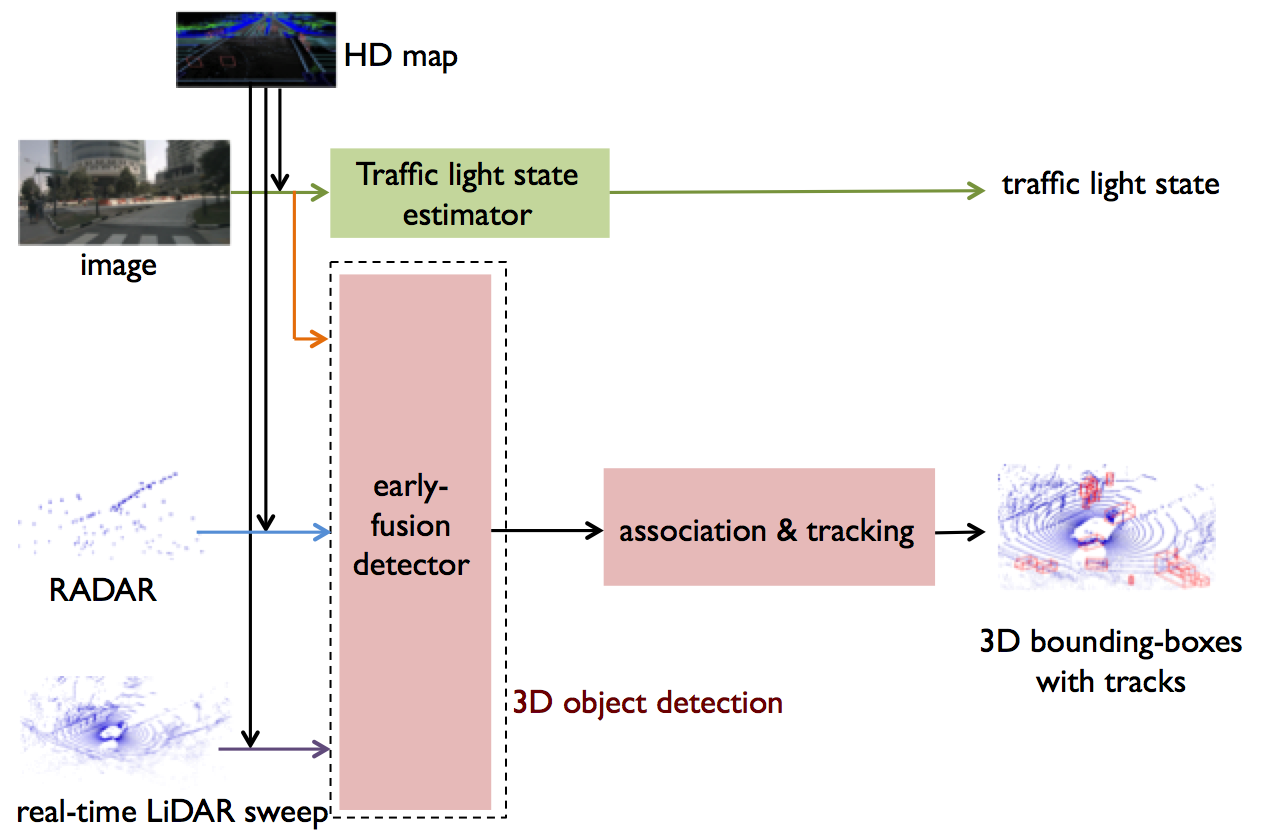}   
    \\
    {\small (a) Late-fusion-based perception module.} & &  {\small  (b) Early-fusion-based perception module.}
  \end{tabular}
\end{center}
\caption{\label{fig:perception} A perception module takes multiple sensing modalities and outputs traffic light states and objects' 3D bounding boxes with tracks. Depending on the mechanism to fuse  modalities, a perception module is categorized into late fusion, which fuses in a semantic space, or early fusion, which fuses in a feature space.}
\end{figure*}

As discussed in Section~\ref{sec:intro_autonomous_system},
multiple sensing modalities are used to ensure the robustness of the perception module. Depending on the mechanism to fuse those modalities, a perception module can be categorized into late fusion and early fusion. Late fusion fuses modalities in a semantic space, which usually happens in  the final step; and early fusion fuses modalities in a feature space, which usually happens in an early or intermediate step. Figure~\ref{fig:perception} (a) shows a standard framework of a late-fusion-based perception module. To obtain objects' 3D bounding boxes with tracks, a late-fusion-based perception module uses an individual pipeline to handle each sensor input. Each pipeline includes the detection component and the association and tracking component. The detection component finds bounding boxes and the association and tracking component tracks bounding boxes across frames to assign a unique identity for each individual object. A late-fusion module unifies the bounding box information from multiple pipelines and outputs a final 3D bounding-boxes with tracks. In comparison, Figure~\ref{fig:perception} (b) shows an early-fusion-based perception module. It uses an early-fusion detector to take the outputs from all the sensing modalities and produce all the 3D bounding boxes. It then uses an association and tracking component to associate 3D bounding boxes across frames and assign an identity for each object.
To estimate traffic light states, a traffic light state estimator extracts the traffic light regions from images according to the position priors in an HD map and then it uses machine learning techniques to analyze the image and identify the traffic light state.

The late-fusion-based approach is much more mature while the early-fusion-based approach is believed to have a bigger potential~\cite{ChenMWLX:17}. The industry has adopted the late-fusion-based approach for decades because this approach modularizes the tasks and makes each sensor pipeline easy to implement, debug and manage. The early-fusion-based approach carries the spirit of end-to-end learning and enables the mutual promotion of multiple sensing modalities in a high-dimensional feature space; however, there are still significant challenges in this research direction and many companies still use the late-fusion-based approach.

A robust perception module usually includes multiple intermediate components, such as lane detection, 2D object detection, 3D object detection, semantic segmentation and object tracking, to achieve the final goal. Among those components, 3D object detection is particularly interesting and challenging because it needs to handle real-time LiDAR sweeps and can directly produce the 3D bounding boxes for all objects in the scene. This task has drawn much attention recently when combined with the power of deep learning~\cite{ChenMWLX:17}. We next focus on 3D object detection.

\subsection{3D object detection} 
The task of 3D object detection is to detect and localize objects in the 3D space with the representation of bounding boxes based on one or multiple sensor measurements. 3D object detection usually outputs 3D bounding boxes of objects, which are the inputs for the component of object association and tracking. Based on the usage of sensor measurements, we can categorize 3D object detection into LiDAR-based detection (see Figure~\ref{sec:perception}(a)) and fusion-based detection (see Figure~\ref{sec:perception}(b)). Qualitative performances are illustrated in Appendix.

\mypar{LiDAR-based detection}
Let $\S$ be a real-time LiDAR sweep.  A LiDAR-based detector aims to find all the objects in the sweep; that is,
\begin{equation}
\label{eq:detection}
\{\oo_i\}_{i=1}^O = h( \S ),
\end{equation}
where $\oo_i = [\y_i, \b_i]$ is the $i$th object in the 3D scene with $\y_i$ the object's category, such as vehicle, bikes and pedestrian, and $\b_i$ the corners of bounding box. Now the detection function $h(\cdot)$ is typically implemented with deep-neural-network-based architectures.

The main difference between 2D object detection and 3D object detection is the input representation. Different from a 2D image, a real-time LiDAR sweep could be represented in various ways, leading to corresponding operations in subsequent components.  For example, PointRCNN~\cite{ShiWL:18}  adopts  the raw-point-based representation~\eqref{eq:raw_point_representation} and then uses PointNet++ with multi-scale
sampling and grouping to learn
point-wise features; 3D FCN~\cite{Li:17} adopts  the 3D-voxelization-based representation~\eqref{eq:voxel_based_representation} and uses 3D convolutions to learn voxel-wise features; PIXOR~\cite{YangLU:18a} adopts the BEV-based representation~\eqref{eq:bev_based_representation} and then uses 2D convolutions to learn pixel-wise features; and FVNet~\cite{ZhouLTSDM:19}, VeloFCN~\cite{LiZX:16} and LaserNet~\cite{MeyerLKVW:19} adopt the range-view-based representation~\eqref{eq:range_view_based_representation} and then use 2D convolutions to learn pixel-wise features. Some other methods consider hybrid representations. VoxelNet~\cite{ZhouT:18} proposes a voxel-feature-encoding (VFE) layer  that combines the advantages of both the raw-point-based representation and  the 3D-voxelization-based representation. VFE first groups 3D points according to the 3D voxel they reside in, then uses PointNet to learn point-wise features in each 3D voxel, and finally aggregates point-wise features to obtain voxel-wise feature for each 3D voxel. The benefit of VFE is to convert raw 3D points to the 3D voxelization-based representation and simultaneously learn 3D geometric features in each 3D voxel. After that, VoxelNet uses 3D convolutions to further extract voxel-wise features. Followed by VoxelNet, Part-A2~\cite{ShiWWL:19} and SECOND~\cite{YanML:19} also adopt VFE. Instead of converting raw 3D points to the 3D voxelization-based representation, a recent detection system, PointPillars~\cite{LangVCZYB:18}, converts raw 3D points to a BEV-based representation, where each pixel in the BEV image corresponds to a pillar in the 3D space. PointPillars then learns pillar-wise features with PointNet and uses 2D convolutions to extract global features from the BEV image.

\begin{figure*}
  \begin{center}
\includegraphics[width=1.45\columnwidth]{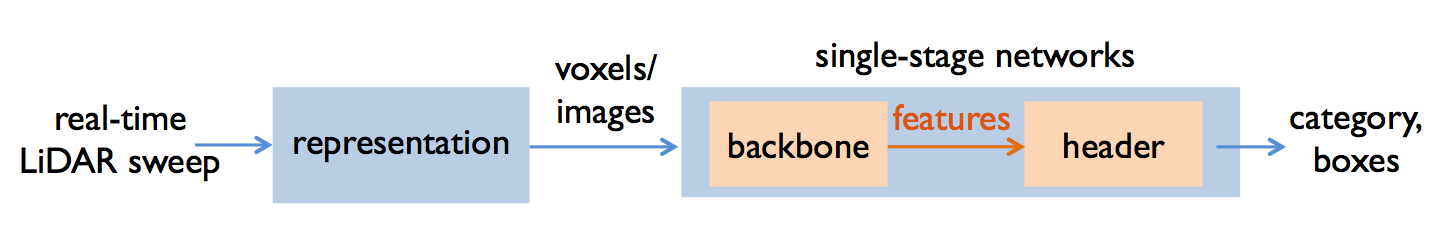}  
  \\
  {\small (a) Single-stage detection framework.}
  \\
\includegraphics[width=1.45\columnwidth]{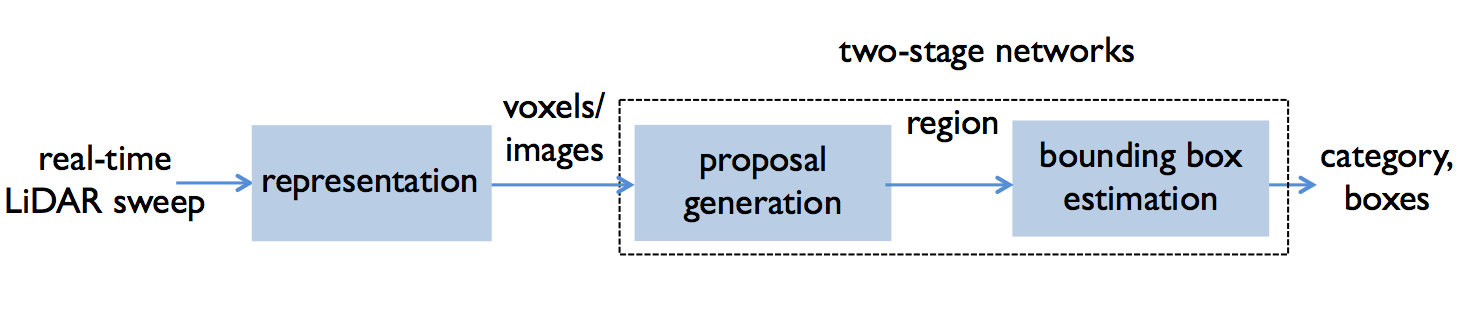}  
\\
{\small (b) Two-stage detection framework.}
  \end{center}
  \caption{\label{fig:detection} The frameworks of the single-stage detection and the two-stage detection. The single-stage detection directly estimates bounding boxes, while the two-stage detection first proposes coarse regions that may include objects and then estimates bounding boxes.}
\end{figure*}

Similarly to 2D objection detection, there are usually two paradigms of 3D object detection:  single-stage detection and two-stage detection; see Figure~\ref{fig:detection}. The single-stage detection directly estimates bounding boxes, while the two-stage detection first proposes coarse regions that may include objects and then estimates bounding boxes.

The single-stage detection directly follows~\eqref{eq:detection}. To implement the detection function $h(\cdot)$, a deep-neural-network architecture  usually includes two components: a backbone, which extracts deep spatial features, and a header, which outputs the estimations. Some methods following the single-stage detection include VeloFCN~\cite{LiZX:16}, 3D FCN~\cite{Li:17}, VoxelNet~\cite{ZhouT:18}, PIXOR~\cite{YangLU:18a}, SECOND~\cite{YanML:19}, PointPillars~\cite{LangVCZYB:18} and LaserNet~\cite{MeyerLKVW:19}. For a backbone, all these methods use 2D/3D convolutional neural networks with multiscale, pyramidal hierarchical structure. Some off-the-shelf  backbone structures are feature pyramid networks~\cite{LinDGHHB:17} and deep layer aggregation~\cite{YuWSD:18}. A header is usually a multitasking network that handles both category classification and bounding box regression. It is usually small and efficient. Some off-the-shelf header structures are single shot detector~\cite{LiuAESRFB:16} and small convolutional neural networks.

The two-stage detection implements the detection function $h(\cdot)$ in two stages; that is,
\begin{subequations}
\label{eq:two_stage_detection}
\begin{eqnarray}
\label{eq:proposal_generation}
\{r_i\}_{i=1}^R & = & h_1( \S ),
\\
\label{eq:bbox_estimation}
\{\oo_i\}_{i=1}^O & = & h_2( \S, \{r_i\}_{i=1}^R ),
\end{eqnarray}
\end{subequations}
where $r_i$ is a set of parameters that describes the $i$th proposed region in the 3D space\footnote{There are multiple approaches to parameterizing a 3D region~\cite{ShiWL:18,ZhouLTSDM:19,ShiWWL:19}.}. The proposal-generation stage~\eqref{eq:proposal_generation} proposes several 3D regions that may include objects inside; and the bounding-box-estimation stage~\eqref{eq:bbox_estimation} extracts 3D points from those proposed regions and estimates the precise object positions.

Some methods following the two-stage detection include PointRCNN~\cite{ShiWL:18}, FVNet~\cite{ZhouLTSDM:19}
and Part-A2 Net~\cite{ShiWWL:19}. In the proposal-generation stage, PointRCNN uses PointNet++ as the backbone and proposes the bin-based localization to propose regions. The bin-based localization first finds the bin associated with the center location of an object and then regresses the residual; Part-A2 Net uses a U-net-like architecture with sparse convolution and deconvolution as the backbone; and FVNet uses feature pyramid networks as the backbone and introduces truncated distances to parameterize a proposed region. In the  bounding-box-estimation stage, both PointRCNN and FVNet use canonical transformation  to align 3D points in each proposed region and PointNet to estimate the parameters of 3D bounding boxes; and Part-A2 Net uses sparse convolutions in a hierarchical way to aggregate features from local to global scales and capture the spatial distribution of proposed regions.

To summarize, the input representation plays a crucial role in the LiDAR-based detection. The raw-point-based representation provides complete point information, but lacks the spatial prior. PointNet has become a standard method to handle this issue and extract features in the raw-point-based representation. The 3D voxelization-based representation and the BEV-based representation are simple and straightforward, but result in a lot of empty voxels and pixels. Feature pyramid networks with sparse convolutions can help address this issue.
The range-view-based representation is more compact because the data is represented in the native frame of the sensor, leading to e.ficient processing, and it naturally models the occlusion. But objects at various ranges would have significantly different scales in the range-view-based representation, it usually requires more training data to achieve high performance. VFE introduces hybrid representations that take advantages of both the raw-point-based representations and the 3D voxelization-based representation. The one-stage detection tends to be faster and simpler, and naturally enjoys a high recall, while the two-stage detection tends to achieve higher precision~\cite{LinGGHD:17,YangLU:18a}. 

\mypar{Fusion-based detection}
A real-time LiDAR sweep provides a high-quality 3D representation of a scene; however, the measurements are generally sparse and only return instantaneous locations, making it difficult for LiDAR-based detection approaches to estimate objects' velocities and detect small objects, such as pedestrians, at range. On the other hand, RADAR directly provides motion information and 2D images provides dense measurements. It is possible to naively merge detections from multiple modalities to improve overall robustness, but the benefit of this approach is limited. Following the end-to-end fashion in deep neural networks, early fusion is believed to be a key technique to significantly improve the detection performance; however, it remains an unresolved problem to design an effective early-fusion mechanism. The main challenges are the following: (1) measurements from each modality come from different measurement spaces. For example, 3D points are sparsely scattered in a continuous 3D space, while images contain dense measurements supported on a 2D lattice; (2) measurements from each modalty are not perfectly synchronized. LiDAR, camera and RADAR capture the scene at their own  sampling frequencies; and (3) different sensing modalities have unique characteristics. The low-level processing of the sensor data depends on the individual sensor modality, but the high-level learning and fusion need to consider the characteristics across multiple modalities.

Some existing early-fusion-based detection systems include MV3D~\cite{ChenMWLX:17}, AVOD~\cite{KuMLHW:18}, F-PointNet~\cite{QiLWSG:18}, PointFusion~\cite{XuAJ:18},  ContinuousConvolution~\cite{LiangYWU:18}, MMF~\cite{LiangYCHU:19} and LaserNet++~\cite{MeyerCHLV:19}. Here we briefly discuss each system.
\begin{itemize}
\item MV3D follows the two-stage detection~\eqref{eq:two_stage_detection} and takes an image and both the BEV-based representation and the range-view-based representation of  a real-time LiDAR sweep as input. MV3D then uses a deep fusion network to enable more interactions among features of the intermediate layers from different views. The fused features are used to jointly classify object categories and estimate 3D bounding boxes; 

\item AVOD follows the two-stage detection~\eqref{eq:two_stage_detection}. It fuses full-resolution feature crops from both the image and the BEV-based representation of a real-time LiDAR sweep;

\item  F-PointNet follows the two-stage detection~\eqref{eq:two_stage_detection}. It extracts the 2D bounding boxes from image detectors and projects to the 3D space to obtain frustum regions. Within each proposed region, F-PointNet uses PointNet to segment 3D instances and estimate 3D bounding boxes;

\item  PointFusion follows the single-stage detection~\eqref{eq:detection}. It first uses convolutional neural networks and PointNet to extract features from an image and the raw-point-based representation of of a real-time LiDAR sweep, respectively. A dense fusion network is then used to combine both features;
that is, for each point, point-wise features are concatenated with image features;

\item  ContinuousConvolution follows the single-stage detection~\eqref{eq:detection} and is based on the BEV-based representation. ContinuousConvolution proposes continuous fusion layers to fuse the image features onto the BEV feature map at various levels of resolution. For each pixel in the BEV feature map, a continuous fusion layer finds its nearest LiDAR point, projects the point onto
the 2D image and retrieves image feature from the corresponding pixel; 
\item  MMF follows the two-stage detection~\eqref{eq:two_stage_detection}. Its fusion mechanism is similar to ContinuousConvolution. Additionally, it introduces depth completion to promote cross-modality feature representation; and

\item  LaserNet++ follows the single-stage detection~\eqref{eq:two_stage_detection}. Based on the range-view-based representation, LaserNet++ builds a pixel-to-pixel correspondence between a camera image and a range-view-based LiDAR image, which allows the algorithm to fuse features extracted from the camera image with the features from the corresponding position of the LiDAR image. It then feeds the features extracted from both the camera image and the LiDAR image to LaserNet~\cite{MeyerLKVW:19}.
\end{itemize}
Each of these works has  shown that adding image data can improve detection performance, especially when LiDAR data is sparse; however, the benefit is not substantial and there is no consensus on a system prototype or a basic operation. This makes the industry hard to overturn the previous late-fusion-based approaches.

To summarize, it remains an open problem to design an early-fusion-based detection system.
Most designs are based on concatenation of intermediate features from both images and 3D point clouds, allowing the networks to figure out how to merge them. So far, there has been no specific design to handle the unsynchronization issue of multiple sensors, which might be implicitly handled by learning from large-scale training data.

\mypar{Datasets}
High-quality datasets are required to train any of the referenced machine learning models. KITTI~\cite{GeigerLU:12} is the most commonly used autonomous-driving dataset, which was released in 2012 and has been updated several times since then. Most 3D object detection algorithms are validated on KITTI; however, KITTI is a relatively small dataset and does not provide detailed  map information. Several autonomous-driving companies have recently released their datasets, such as nuScenes\footnote{\url{https://www.nuscenes.org/}}, Argoverse\footnote{\url{https://www.argoverse.org/}},  Lyft Level 5 AV dataset\footnote{\url{https://level5.lyft.com/dataset/}} and the Waymo open dataset\footnote{\url{https://waymo.com/open}}.

\mypar{Evaluation metrics}
To evaluate the detection performance, standard evaluation metrics in academia are the precision-recall (PR) curve and average precision (AP); however, there is no standard platform to evaluate the running speed of each model. On the other hand, industry considers more detailed evaluation metrics to check the detection performances. For example, practitioners would check the performances at various ranges, shapes, sizes, appearances, and occlusion levels to get more signals. They would also check the influences on the subsequent modules, such as object tracking, future trajectory prediction, and motion planning to obtain the system-level metrics.

\subsection{Real-world challenges}
With the growth of deep learning, the perception module has achieved tremendous improvements. Some practitioners no longer consider it as the technical bottleneck of autonomous driving; however, the perception module is still far from perfect. Here are a series of challenges in the perception module.

\mypar{High cost} A self-driving vehicle is usually equipped with one or more LiDARs and computing devices, such as GPUs and other specialized processors, which are expensive. The high cost makes it formidable to maintain a scaled fleet of autonomous vehicles. It remains an open problem to exploit information from real-time LiDAR sweeps using low-cost computation;

\mypar{Tradeoffs between effectiveness and efficiency} A self-driving vehicle should react to its surroundings in real-time. It would be  meaningless to  pursue a high-precision perception module when it introduces too much latency; however, researchers tend to focus much more on the effectiveness than the efficiency of an algorithm;

\mypar{Training data deluge} A modern perception module  heavily  depends on machine learning techniques, which usually need as much training data as possible; however, it takes a lot of time and computational resources to handle large-scale training data. It remains a yet to be resolved problem to effectively choose a representative subset of training data from the entire dataset, which would significantly accelerate the product development;

\mypar{Long-tail issues} There are countless traffic conditions where large-scale training data cannot cover all the possibilities. It remains an unresolved problem to find and handle corner cases, especially detecting objects that never appear in the training data;

\mypar{Research conversion} In academia, research tends to design algorithms based on clean and small-scale  datasets. It turns out that many effective algorithms work well for those clean and small-scale datasets, but are ineffective on noisy and large-scale datasets. Meanwhile, some algorithms that work well on large-scale datasets do not work well on small-scale datasets~\cite{MeyerLKVW:19}. These discrepencies can reduce the usefulness of research results when applied to real-world problems. Industry should consider providing representative datasets and perhaps even a computational evaluation platform that allows people to compare various methods at full industrial scale; and

\mypar{Evaluation metrics} Objects in a scene have various levels of interactions with an autonomous vehicle. Incorrect estimations of some objects would lead to much bigger consequences than that of other objects; however, the PR curve and AP give uniform weights to all the samples. Additionally, the PR curve and AP do not clearly reflect corner cases, which have only a small sample size; Thus, improving the PR curve and AP do not necessarily lead to a better behavior of an autonomous vehicle. It is often more important to slice the test data and look at the performance over subsets of high-impact cases in addition to overall AP. A standardized simulator could also be developed to provide some system-level metrics.

\section{Summary and Open Issues} 
\label{sec:perspectives}
The field of autonomous driving is experiencing rapid growth. Many techniques have become relatively mature; however, an ultimate solution for autonomous driving has yet to be determined. At the current stage, LiDAR is an indispensable sensor for building a reliable autonomous vehicle, and advanced techniques for 3D point cloud processing and learning are critical building blocks for autonomous driving. In this article, we surveyed recent developments in the area of 3D point cloud processing and learning and presented their applications to autonomous driving.  We described how 3D point cloud processing and learning makes a difference in three important modules in autonomous driving: map creation, localization and perception.

With the rapid development of 3D point cloud processing and learning, the overall performances of the map creation, localization and perception modules in an autonomous system have been significantly improved; however, quite a few challenges remain ahead. Here we briefly mention a few important open issues.

\textbf{How should we make processing and learning algorithms scalable and efficient?} Now we are still in the developing phase and autonomous vehicles are tested in a limited number of canonical routes or over a small area. In the near future, autonomous vehicles might be tested in a city/country scale, which needs a city/country-scale HD map. This requires scalable algorithms to create and update HD maps. Now an autonomous vehicle is usually equipped with a 64-line LiDAR, which still produces relatively sparse point clouds. In the near future, LiDAR might have many more lines and produce much denser point clouds. This requires more efficient algorithms to achieve LiDAR-to-map localization and 3D object detection in the real-time;
    
\textbf{How should we make processing and learning algorithms robust enough to handle corner cases?} We can collect large amounts of real-world sensor data and generate large amounts of simulated sensor data, but we need to deliberately select the most representative data to improve the generality of the algorithms. At the same time, one has to face the fact that all learning algorithms depend on training data, which can never cover all the possibilities. To address this issue, one key research area  is to improve the uncertainty estimation of an algorithm, because this allows a system to react conservatively when the learned components are not confident. This requires reasoning both about the known uncertainty from the training data and also the more challenging uncertainty from cases that are not covered by the training data;
    
\textbf{How should we develop processing and learning algorithms with a faster iteration speed?} We want more data and more complicated algorithms to achieve better performance for autonomous driving; meanwhile, we want efficient and practical algorithms to accelerate product development, which is also critical. Practitioners in industry should collaborate closely with researchers in academia to increase the research conversion rate; and

\textbf{How should we evaluate processing and learning algorithms?} Currently most processing and learning algorithms are evaluated on specific model-level metrics to meet the criteria of the corresponding tasks; however, these model-level metrics often do not fully correlate with system-level metrics that reflect the overall behavior. Along these same lines, the research community often focuses on improving the average performance, but there needs to be an increased focus on improving the rare long-tail cases that are really critical for a real-world system.

Note that adversarial attack is a potential issue; however, it is not one of the most critical challenges because the current techniques are far away from the performance level where adversarial attack could be a major concern.

\clearpage
\appendix
\section{Appendices}
\subsection{Relations between academia and industry} 
In terms of studying 3D point cloud processing and learning, we compare academia and industry from four aspects: specific aim, dataset, methodology and evaluation metrics.

\mypar{Specific aim} Researchers in academia generally abstract a real-world problem to a specific, standardized setting, often with a fixed dataset and metric to optimize. They then focus on this setting, propose algorithms and make comparisons, pushing forward the state of the art by showing improvements within this setting. On the other hand, practitioners in industry generally focus on system-level tasks and what is required to make a system work in a real-world setting. This often includes complex system dependencies, multiple metrics to satisfy, and datasets that grow over time. They push forward the state of the art by showing a system that performs well in the real-world application. For example, to create an HD map, researchers abstract a high-level research problem: 3D point cloud registration.  To achieve this, they propose a classical registration algorithm, iterative closest point (ICP)~\cite{BeslM:92}. Based on the formalism of this algorithm, some researchers study its theoretical properties, such as convergence; other researchers extend it to various advanced versions, such as point-to-plane ICP~\cite{Low:04} and global ICP~\cite{YangLCJ:16}, to tackle various specific settings. To solve the same map creation task, practitioners would combine an ICP based registration process with additional sensor data from a GPS and IMU to develop a more robust system that can operate effectively and efficiently on real-world problems.

\mypar{Datasets} Researchers in academia work with small-scale, specific datasets, while practitioners in industry have to use large-scale, noisy, comprehensive datasets. For example, to detect 3D bounding boxes in the perception module, researchers use the KITTI dataset~\cite{GeigerLU:12,GeigerLSU:2013}, which has only a few thousands LiDAR sweeps; to recognize 3D point clouds, researchers use the ModelNet 40 dataset~\cite{WuSKYZTX:15}, which has only a few thousands models. A small-scale dataset eases the computational cost and makes it fast to iterate on the algorithms. To solve the same detection task, practitioners would use much bigger datasets to make the model more robust and handle the long-tail phenomenon. On the other hand, the research community in academia is larger, the datasets are smaller, and the problem is more focused.
Therefore, academia can generally iterate faster than industry. But sometimes, academia might make the wrong conclusions due to overfitting problems to a small dataset (i.e. KITTI) or early discarding of more powerful methods that require more data to generalize and converge.

\mypar{Methodology} Researchers in academia emphasize the technical novelty, while practitioners in industry consider the trade-off between effectiveness and efficiency and focus on practical solutions to real-world problems. For example, to localize an autonomous vehicle, researchers may consider various approaches based on SLAM, which is technically interesting; however, practitioners would prefer using an offline HD map, which demands expensive resources to build the map but the map-based localization can be highly efficient and robust compared to SLAM; and

\mypar{Evaluation metrics} Researchers in academia use focused model-level metrics, while practitioners in industry generally use a large number of model-level and system-level evaluation metrics to ensure the robustness of the proposed algorithms. For example, to detect 3D bounding boxes in the perception module, researchers usually use the precision-recall (PR) curve and average precision (AP) to judge a detection algorithm, which is easy to make comparisons in a research paper; however, practitioners would propose various metrics to gain more insights to the algorithm. Instead of relying solely on the overall PR curve and AP, they would check the performances at various range categories and the influences on the subsequent modules to understand the overall system performance.

\subsection{Qualitative results}
\begin{figure*}[htb]
  \begin{center}
    \begin{tabular}{cc}
   \includegraphics[width=0.7\columnwidth]{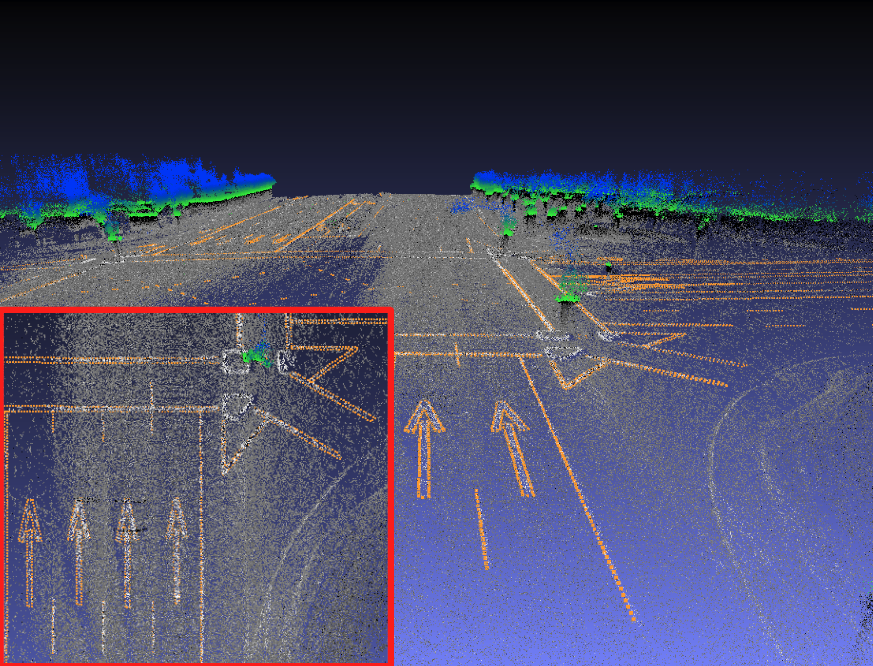}   
    & \includegraphics[width=0.8\columnwidth]{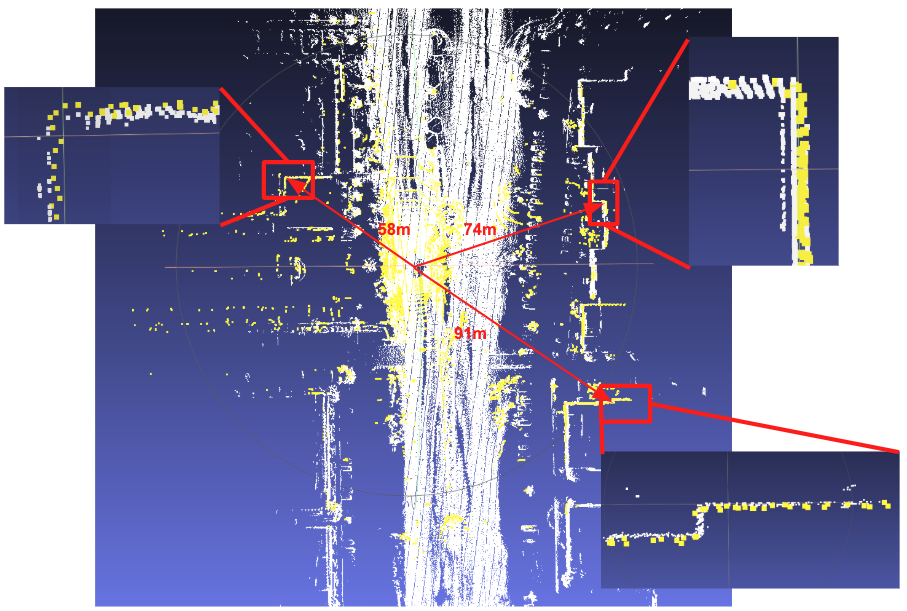} 
  \end{tabular}
\end{center}
\caption{\label{fig:hdmaps_snapshots} Illustration of high-definition maps and real-time localization.  Plot (a) shows a sample portion of an HD map, which includes both a point-cloud map and traffic-rule-related semantic feature map. Plot (b) shows that a real-time LiDAR sweep (yellow point cloud) matches with the point-cloud map (white point cloud).}
\end{figure*}

To illustrate the high-definition maps and real-time map-based localization, we present Figure~\ref{fig:hdmaps_snapshots} with permission from Precivision Technologies, Inc. Figure~\ref{fig:hdmaps_snapshots}(a) shows a sample portion of HD maps in Santa Clara, CA, USA. In this figure, the 3D contours of the lane marker features (shown in orange colar) are overlaid onto the point-cloud map, where the ground color of the point-cloud map represents the laser reflectivity (white color indicates high reflectivity, black color indicates low reflectivity); the blue-to-green color of the point cloud represents the height of the point. Note that the high laser reflectivity points (i.e. the white color points) in the ground point cloud are the points on lane markers, and their sharp boundaries qualitatively demonstrate the centimeter-level local precision. 
Figure~\ref{fig:hdmaps_snapshots}(b) shows the bird's eye view visualization of the registration between a real-time LiDAR sweep and a point-cloud map. In this figure, yellow point cloud indicates a LiDAR sweep, and white point cloud indicates the point-cloud map. Note that good alignment between the real-time LiDAR sweep and the point-cloud map is demonstrated through three examples of zoomed details (as shown in the insets of this figure), where these details are chosen as the portions $>50$m away from the position of the LiDAR. Both the centimeter level precision for the translation component and the micro-radian level precision for the rotation component of the alignment (i.e. $10$ centimeters $/$ $50$m = $2$ mrad) are qualitatively demonstrated.

\begin{figure*}[thb]
  \begin{center}
  \includegraphics[width=1.5\columnwidth]{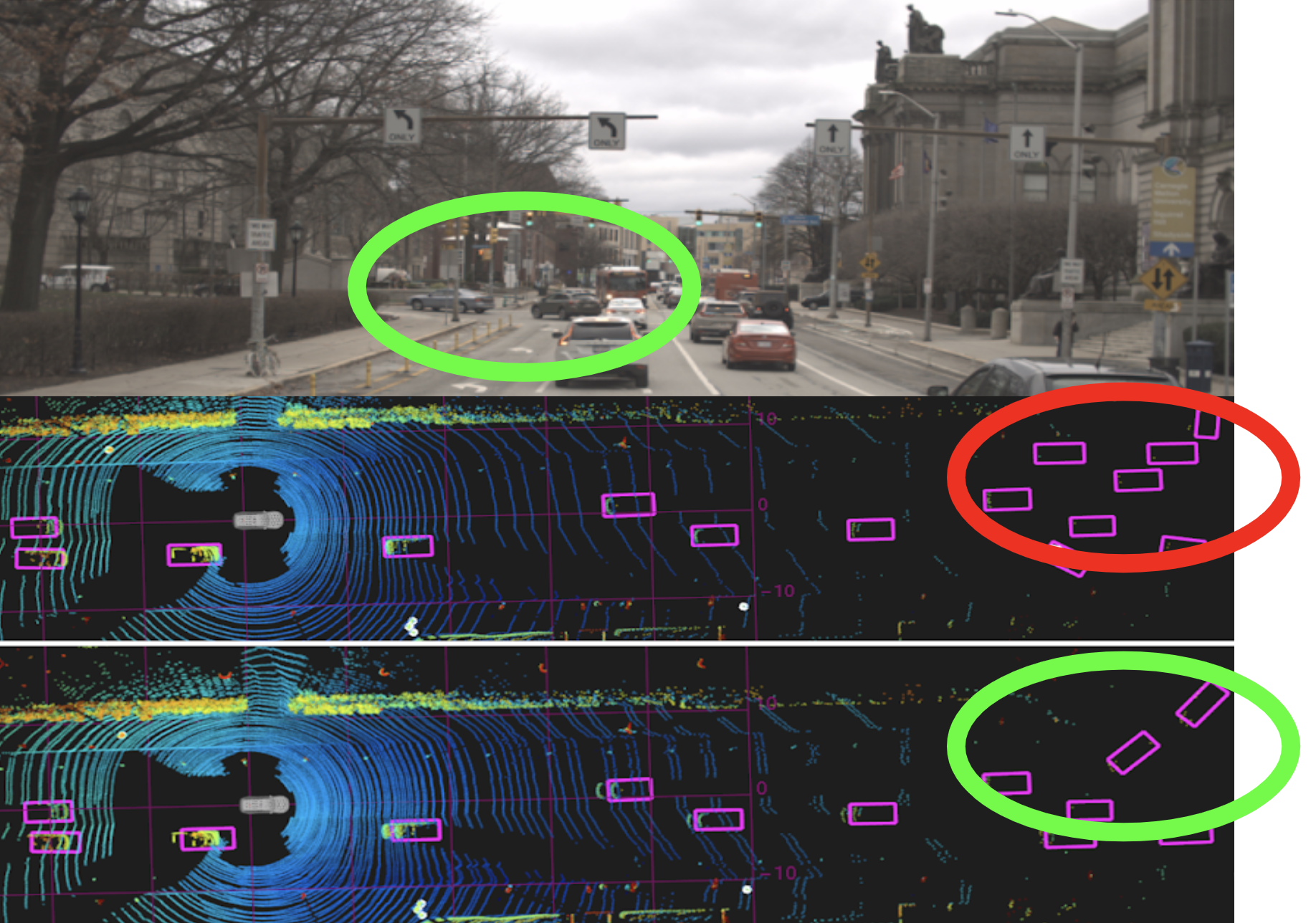} 
\end{center}
\caption{\label{fig:fusion_results} Comparison between LaserNet~\cite{MeyerLKVW:19}, a LiDAR-based detector, and LaserNet++~\cite{MeyerCHLV:19}, a fusion-based detector (LiDAR + image). The middle row shows the bird's eye view visualization of the output of LaserNet and the botton row shows the bird's eye view visualization of the output of LaserNet++. We see that LaserNet++ outperforms LaserNet especially when the objects are far away from the autonomous vehicle. The differences are highlighted in the red and green circles.}
\end{figure*}

To illustrate 3D object detection, we present Figure~\ref{fig:fusion_results}, which is from~\cite{MeyerCHLV:19} with permission. The model is called LaserNet++, which is a state-of-the-art fusion-based 3D object detector and developed at Uber Advanced Technologies Group. LaserNet++ takes both LiDAR and camera data as input and is trained on a dataset containing 5,000 sequences sampled at 10 Hz for a total of 1.2 million images. In Figure~\ref{fig:fusion_results}, we compare LaserNet++ with a state-of-the-art LiDAR-based 3D object detector, LaserNet~\cite{MeyerLKVW:19}, which is also developed at Uber Advanced Technologies Group. The middle row shows the bird's eye view visualization of the output of LaserNet and the botton row shows the bird's eye view visualization of the output of LaserNet++. We see that LaserNet++ outperforms LaserNet especially when the objects are far away from the autonomous vehicle. This indicates the importance of fusing information from multiple modalities.

\subsection{Elementary tasks}
\label{sec:tasks}
There are increasing needs for 3D point clouds in various application scenarios, including autonomous systems~\cite{ZhouT:18}, robotics systems~\cite{PomerleauCS:15}, infrastructure inspection~\cite{FathiDL:15}, virtual and augmented reality~\cite{ReitingerZS:07}, animation~\cite{SussmuthWG:08} and preservation of historical artifacts~\cite{LozesEL:15}.  3D point cloud processing and learning naturally extend numerous tasks in 1D signal processing, 2D image processing, machine learning and computer vision to the 3D domain. In this section, we consider a few representative tasks that have received great attention in academia: learning tasks, such as reconstruction, recognition and segmentation, as well as processing tasks, such as denoising, downsampling, upsampling and registration. Those elementary tasks abstract real-world problems to simplified and standardized settings and work as testbeds for developing new tools, which can be potentially applied to the map creation, localization and perception modules in a real-world autonomous system.

\subsubsection{3D point cloud reconstruction}
\begin{figure}
  \begin{center}
\includegraphics[width=0.9\columnwidth]{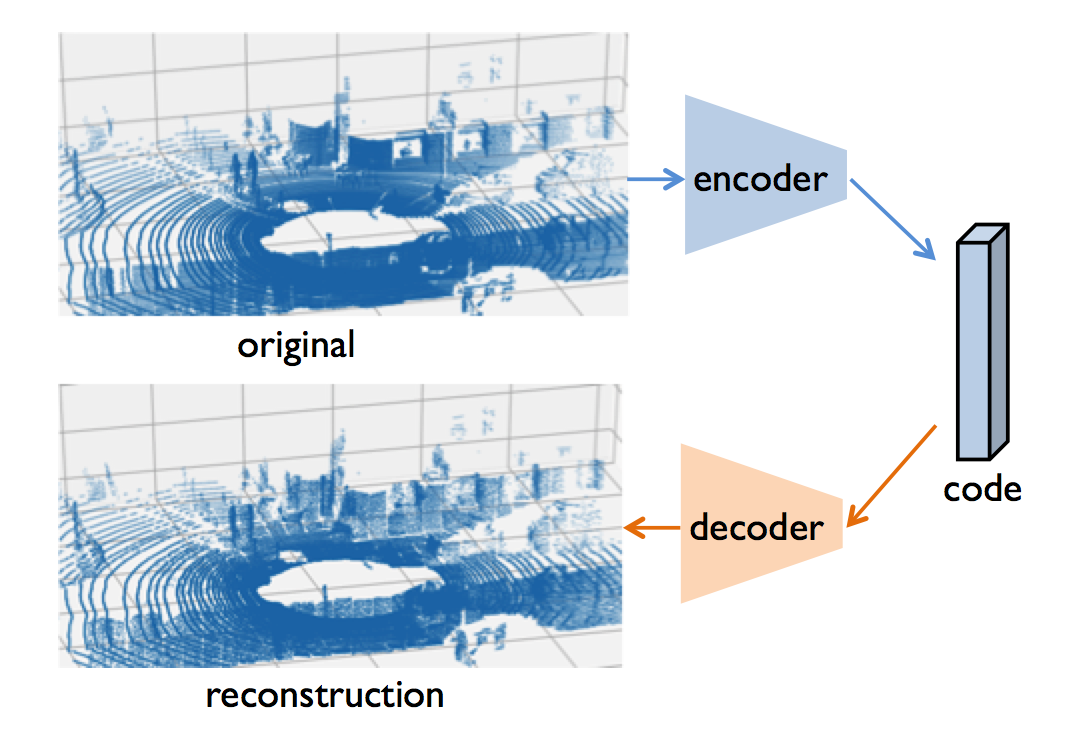} 
  \end{center}
  \caption{\label{fig:reconstruction}  3D point cloud reconstruction aims to find  a compact representation of a 3D point cloud that preserves the ability to reconstruct the original 3D point cloud.}
\end{figure}

The goal is to find  a compact representation of a 3D point cloud that preserves the ability to reconstruct the original 3D point cloud; see Figure~\ref{fig:reconstruction}.  Reconstruction is helpful for data storage in autonomous driving. Since each autonomous vehicle needs to store an HD map and collect real-time LiDAR sweeps, data storage would be expensive for a large fleet of autonomous vehicles. Although there is no mature compression standard to handle large-scale, open-scene 3D point clouds~\cite{SchwarzPBBCCCKL:19}, reconstruction techniques could provide 3D point cloud compression and reduce the cost for data storage in autonomous driving.

Here we only consider the 3D coordinate of each 3D point. Let $\S = \{\x_i \}_{i=1}^N$ be a set of $N$ 3D points, whose $i$th element $\x_i \in \R^3$ is the 3D coordinate. We aim to design a pair of an encoder $\Psi(\cdot)$ and a decoder $\Phi(\cdot)$, such that $\Psi(\cdot)$ compresses a 3D point cloud $\S$ to a low-dimensional code $\mathbf{c}$ and $\Phi(\cdot)$  that decompress the code back to a reconstruction $\widehat{\S} = \{\widehat{\x}_i \}_{i=1}^M$ that approximates $\S$; that is,
\begin{subequations}
\label{eq:reconstruction}
\begin{eqnarray}
\label{eq:encoder}
\mathbf{c}  & = &  \Psi \left( \S \right) \in \R^{C}
\\
\label{eq:decoder}
 \widehat{\S}  & = & \Phi \left( \mathbf{c}  \right),
\end{eqnarray}
\end{subequations}
where the code $\mathbf{c}$ with $C \ll 3N$ summarizes the original point cloud. We aim to optimize over  $\Psi(\cdot)$ and $\Phi(\cdot)$ to push $\widehat{\S}$ to be close to $\S$. Note that the number of points in the reconstruction may be different from the number of points in the original 3D point cloud.

To evaluate the quality of reconstruction, two 
distance metrics are usually considered. The Earth mover's distance is the objective function of a transportation problem, which moves one point set to the other with the lowest cost; that is,
\begin{eqnarray}
  \label{eq:emd}
  d_{\rm EMD}(\S, \widehat{\S}) & = &
   \min_{\phi: \S \rightarrow \widehat{\S}} 
   \sum_{\x \in \S}  \left\| \x - \phi(\x)  \right\|_2,
\end{eqnarray}
where $\phi$ is a bijection. The Chamfer distance measures the total distance between each point in one set to its nearest neighbor in the other set; that is,
\begin{eqnarray*}
  \label{eq:chamfer}
  d_{\rm CH}(\S, \widehat{\S}) & = &
    \frac{1}{N} \sum_{\x \in \S}  \min_{ \widehat{\x} \in  \widehat{S}}  \left\|  \x - \widehat{\x} \right\|_2  +  \frac{1}{M} \sum_{\widehat{\x} \in  \widehat{S}}  \min_{\x \in \S}  \left\|   \widehat{\x} - \x \right\|_2  .
\end{eqnarray*}
Both the Earth mover's distance and the Chamfer distance enforce the underlying manifold of the reconstruction to stay close to that of the original point cloud. Reconstruction with the Earth mover's distance usually outperforms that with the Chamfer distance; however, it is more efficient to compute the Chamfer distance.

\mypar{Standard experimental setup} 
A standard dataset is ShapeNet~\cite{ChangFGHHLSSSSX:15}. It contains more than 3,000,000 3D models, 220,000 models out of which are classified into 3,135 categories (WordNet synsets). For each 3D model, one can sample 3D points from the surfaces of these 3D models by using the Poisson-disk sampling algorithm~\cite{Bridson:07} and rescale the points into a unit cube centered at the origin. The evaluation metric is either the Earth mover's distance or the Chamfer distance.

\mypar{Standard methods} 
An encoder should extract global features that preserve as much information as possible from an original 3D point cloud. To design an encoder, one can adopt either PointNet-based methods or graph-based methods~\cite{AchlioptasDMG:18, YangFST:18}. For example, latentGAN~\cite{AchlioptasDMG:18} directly uses the global feature vector $\h$~\eqref{eq:PointNet} from PointNet to encode the overall geometry information of a 3D point cloud.

A decoder should translate information from the feature space to the original 3D space as much as possible. To design a decoder, the simplest approach is to use fully-connected neural networks, which work well for a small-scale 3D point cloud, but require a huge number of training parameters~\cite{AchlioptasDMG:18}. 

To improve efficiency, FoldingNet~\cite{YangFST:18} and AtlasNet~\cite{GroueixFKRA:18} consider the decoding as a warping process that folds a 2D lattice to a 3D surface. Let $\Z \in \mathbb{Z}^{M \times 2}$ be a matrix representation of  nodes sampled uniformly from a fixed regular 2D lattice and the $i$th row vector $\z_i \in \R^2$ be the 2D coordinate of the $i$th node in the 2D lattice. Note that $\Z$ is fixed and is used as a canonical base for the reconstruction, which does not depend on an original 3D point cloud.  We  can then concatenate each 2D coordinate with the code from the encoder~\eqref{eq:encoder} to obtain a local feature; and then, use MLPs to implement the warping process. Mathematically, the $i$th point after warping is
\begin{equation*}
\label{eq:folding}
 \widehat{\x}_i  \ = \  g_{\mathbf{c} }( \z_i )  =  {\rm MLP} \left( \left[   {\rm MLP} \left( \left[ \z_i,  \mathbf{c}  \right]  \right),  \mathbf{c}  \right]  \right) \in \mathbb{R}^{3},
\end{equation*}
where the code $\mathbf{c}$ is the output of the encoder and $\left[  \cdot, \cdot \right] $ denotes the concatenation of two vectors. 
The warping function $g_{\mathbf{c} }( \cdot )$ consists of two-layer MLPs and the code is introduced in each layer to guide the warping process. We collect all the 3D points $\widehat{\x}_i$ to form the reconstruction $\widehat{\S} = \{ \widehat{\x}_i \in \R^3, i= 1, \cdots, M \}$. 

Intuitively, introducing a 2D lattice provides a smoothness prior; in other words, when two points are close in the 2D lattice, their correspondence after warping are also close in the 3D space.  This design makes the networks easy to train and save a huge amount of training parameters.

Most 3D reconstruction algorithms consider small-large 3D point clouds, representing individual objects; however, there could be huge variations of 3D shapes in large-scale, open scenarios. To make the reconstruction algorithms practical for large-scale 3D point clouds, the point cloud neural transform (PCT)~\cite{ChenNLL:19} combines voxelization and learning. We can discretize the 3D space into nonoverlapping voxels and then use the neural networks to compactly represent 3D points in each voxel. This voxel-level representation not only introduces more training samples, but also reduces the possibility of shape variations.

\subsubsection{3D point cloud recognition}
\label{sec:recognition}
\begin{figure}
  \begin{center}
\includegraphics[width=0.9\columnwidth]{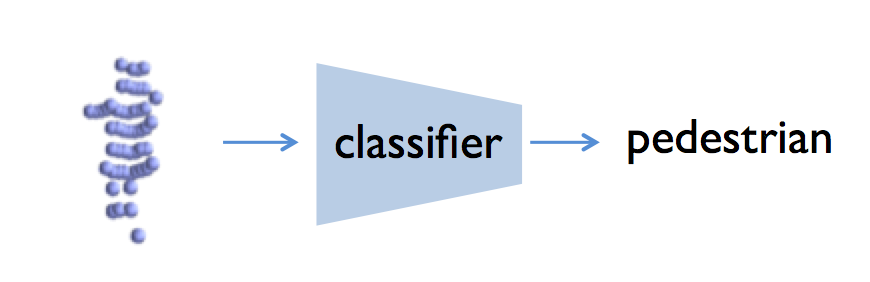} 
  \end{center}
  \caption{\label{fig:recognition}  3D point cloud recognition aims to classify a 3D point cloud to a predefined category. }
\end{figure}

The goal of recognition is to classify a 3D point cloud to a predefined category; see Figure~\ref{fig:recognition}. As a typical task of 3D point cloud learning, recognition is critical to the perception module in autonomous driving, where we aim to classify objects in the 3D scene.
 
Let $h$ be a classifier that maps a 3D point cloud $\S$ to a confidence vector $\y$ indicating the category belongings, that is, 
\begin{equation}
 \y = h(\S) \in [0, 1]^C,
\end{equation}
where $C$ is the number of classes. The $c$th element of $\y$, $\y_c$, indicates the likelihood of the 3D point cloud belonging to the $c$th class. 

\mypar{Standard experimental setup}  A standard dataset is ModelNet40~\cite{WuSKYZTX:15}. It contains 12,311 meshed CAD models from 40 categories. A standard validation paradigm is to use $9,843$ models for training and $2,468$ models for testing. For each CAD model, one can  sample 3D points uniformly from the mesh faces; the 3D point cloud is rescaled to fit into the unit sphere. The evaluation metric is the classification accuracy.

\mypar{Standard methods} 
As with many other classification tasks, 3D point cloud recognition used to involve two phases: the feature-extraction phase and the classification phase. After the emergence of deep neural networks, the state-of-the-art algorithms for 3D point cloud recognition have been based on  end-to-end supervised neural network architectures. Recently, many researchers have made a lot of efforts in pursuing PointNet-based methods and graph-based methods. Based on a softmax layer, both approaches can be used to train an end-to-end classifier in a supervised fashion and achieve strong classification performances.

Additionally, unsupervised-learning (reconstruction) methods can be used for supervised tasks. As discussed in~\eqref{eq:reconstruction}, we can train an encoder-decoder network and then use the code obtained from the encoder to train a classifier and achieve 3D point cloud recognition. This method involves two standard phases: the training of the encoder-decoder network and the training of the classifier. Each phase uses an individual dataset. Recent works in this area include FoldingNet~\cite{YangFST:18}, AtlasNet~\cite{GroueixFKRA:18} and many others~\cite{ChenDYLFT:19}. This approach usually performs worse than the end-to-end supervised-learning architectures, but it has a better generalization ability~\cite{YangFST:18}. The reason is that hidden features of unsupervised learning are not directly trained based on the final labels.

\subsubsection{3D point cloud segmentation}
\label{sec:segmentation}
\begin{figure}
  \begin{center}
\includegraphics[width=0.9\columnwidth]{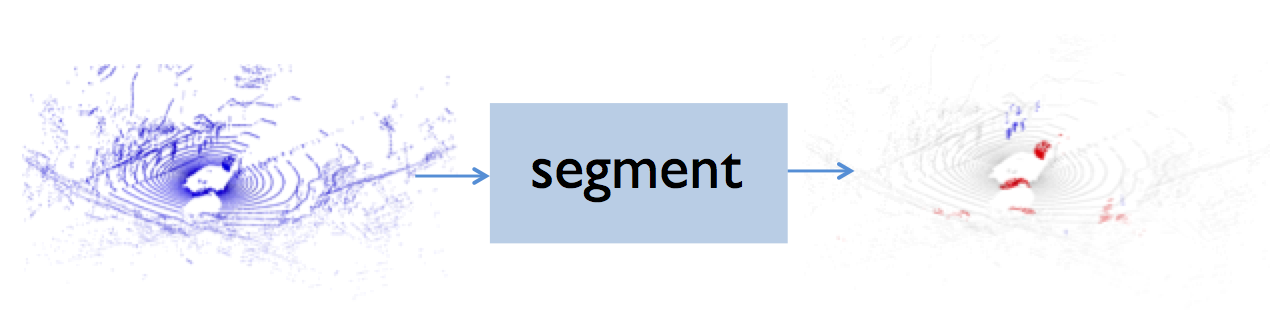} 
  \end{center}
  \caption{\label{fig:segmentation}  3D point cloud segmentation aims to classify each 3D point in a 3D point cloud to a predefined category. }
\end{figure}

The goal of segmentation is to classify each 3D point in a 3D point cloud to a predefined category; see Figure~\ref{fig:segmentation}. As a typical task of 3D point cloud learning, segmentation is critical in the perception module of autonomous driving, where we want to identify the class of each LiDAR point, such as vehicle, tree or road. Let $h$ be the classifier that maps each 3D point in a 3D point cloud $\x_i \in \S$ to a confidence vector $\y$, that is, 
\begin{equation}
 \y_i = h(\x_i) \in [0, 1]^C,
\end{equation}
where $C$ is the number of classes. The $c$th element of $\y$, ${\y_i}_c$, indicates the probability of the $i$th 3D point belonging to the $c$th class. 

Depending on the scale and semantic meaning of a 3D point cloud, a segmentation task can be categorized to part segmentation and scene segmentation. Part segmentation is to segment an individual object into several parts; while scene segmentation is to segment a large-scale scene into several objects.

\mypar{Standard experimental setup} 
For part segmentation, a standard dataset is ShapeNet part dataset~\cite{YiKCSYSLHSG:16}. It contains $16,881$ 3D models from 16 object categories, in total annotated with $50$ parts. Most 3D models are labeled with less than $6$ parts. 3D points can be sampled from each 3D model. The evaluation metric is the mean  intersection over union (IoU); that is, the ratio between the intersection and the union. Mathematically, 
\begin{equation}
\label{eq:miou}
{\rm mean~IoU} = \sum_{i=1}^{C} \frac{|\widehat{\S}_i \cap \S_i|}{|\widehat{\S}_i \cup \S_i|},
\end{equation}
where $C$ is the number of classes, $\S_i$ and $\widehat{\S}_i$ are the ground-truth and the predicted point sets of the $i$th class, respectively.

For scene segmentation, a standard dataset is Stanford Large-Scale 3D Indoor Spaces Dataset (S3DIS)~\cite{ArmeniSZS:17}. It includes 3D scan point clouds for 6 indoor areas including 272 rooms in total. Each 3D point belongs to one of 13 semantic categories, such as board, bookcase, chair, ceiling, etc., plus clutter. Each 3D point is represented as a 9D vector (XYZ, RGB, and normalized spatial coordinates). The evaluation metrics include the mean IoU~\eqref{eq:miou} and the per-point classification accuracy.

\mypar{Standard methods} The methods used for recognition could be extended to segmentation. For example, In the recognition task, we extract local features for each 3D point and then aggregate all the local features to obtain a global feature for a 3D point cloud. In the segmentation task, one can concatenate local features for each 3D point with global features to obtain final point-wise features; and then use these point-wise features to classify each 3D point. For example, to use PointNet in the segmentation task, we concatenate local features $\Hh_i$ with global features $\h$ to obtain the final point-wise features $\h_{\rm seg} = \left[  \Hh_i， \h \right]$, which includes both local and global information. We then input $\h_{\rm seg}$ to a classifier, such as softmax, to classify the corresponding 3D point.

\subsubsection{3D point cloud denoising}
\begin{figure}
  \begin{center}
\includegraphics[width=0.9\columnwidth]{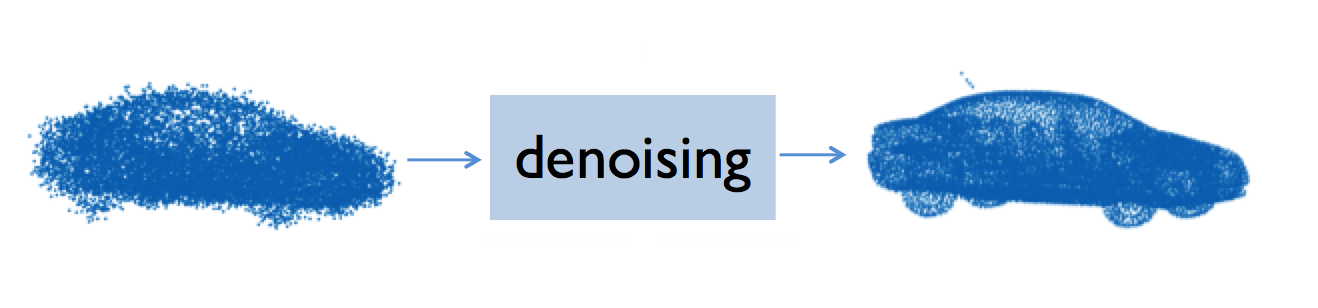} 
  \end{center}
  \caption{\label{fig:denoising}  3D point cloud denoising aims to remove noise from a noisy 3D point clouds and recover an original 3D point clouds. }
\end{figure}

The goal of denoising is to remove noise from a noisy 3D point clouds and recover an original 3D point clouds; see Figure~\ref{fig:denoising}. As a typical task in 3D point cloud processing, denoising is a stealth technique in autonomous driving. In many modules, we want to smooth data for further processing. For example, image-based 3D reconstruction may fail to manage matching ambiguities, leading to noisy 3D point clouds.

Here we only consider the 3D coordinate of each 3D point. Let $\S = \{\x_i \}_{i=1}^N$ be a noiseless 3D point cloud, whose $i$th element $\x_i \in \R^3$ is the 3D coordinate, and $\S^{(\epsilon)} = \{ \x^{(\epsilon)}_i \}_{i=1}^N$ be a noisy 3D point cloud, where $\x^{(\epsilon)}_i = \x_i + \e_i \in  \R^3 $ with Gaussian noise $\e_i \in \R^3$. Let $h$ be a denoiser that maps a noisy 3D point cloud $\S^{(\epsilon)}$ to the original 3D point cloud $\S$, that is,
\begin{equation*}
\widehat{\S} = h(\S^{(\epsilon)}).
\end{equation*}
We aim to optimize over  the denoiser $h(\cdot)$ to push $\widehat{\S}$ be close to $\S$. To evaluate the quality of denoising, mean square errors are usually considered; that is, 
\begin{equation}
  d (\S, \widehat{\S}) \ = \ \sum_{\x_i \in  \S, \widehat{\x}_i \in  \widehat{\S}} \left\|  \x_i - \widehat{\x}_i \right\|_2^2 ,
\end{equation}

\mypar{Standard experimental setup} As a processing task, there is no standard dataset for denoising. Researchers usually use a few toy examples, such as Bunny~\cite{TurkL:94}, to validate denoising algorithms.

\mypar{Standard methods} 
Here we mainly consider two approaches: the filtering-based approach and the optimization-based approach. The filtering-based approach usually runs through a 3D point cloud point by point, replacing the 3D coordinate of each 3D point with weighted average of its neighboring points. In comparison, the optimization-based approach usually introduces a regularization term that promotes
smoothness and solves a regularized optimization problem to obtain a global solution. The optimization-based approach usually outperforms the filtering-based approach, but at the same time, the optimization-based approach may smooth out points around edges and contours.

One representative of the filtering-based approach is bilateral filtering~\cite{KornprobstTD:09}. It is a classical image denoising algorithm that uses a non-linear, edge-preserving, smoothing filter to reduce noises. To adapt it to a 3D point cloud, one can construct a mesh from 3D points. Bilateral filtering then replaces the coordinates of each 3D point with a weighted average of coordinates from nearby 3D points. The neighbors are defined according to the mesh connections. Mathematically, for the $i$th 3D point, bilateral filtering works as
\begin{equation*}
  \widehat{\x}_i \ = \ \frac{\sum_{j \in \N_i} w_{i,j} \x_j}{\sum_{j \in \N_i} w_{i,j}},
\end{equation*}
where $\N_i$ is the neighbors of the $i$th point, $w_{i,j}$ is the weight between the $i$th and the $j$th 3D points, which is flexible to design. A standard choice of the weight is based on a Gaussian kernel; that is, $w_{i,j} = \exp{\left(-\left\|\x_i - \x_j \right\|_2^2 / \sigma^2 \right)}$ with a hyperparameter $\sigma$. The problem of bilateral filtering is over-smoothing.  To solve this issue, a series of works consider dedicated methods to design weights and generalized neighborhood. For example,~\cite{DeschaudG:10} extends the non-local means denoising approach to 3D point clouds and adaptively filter 3D points in an edge preserving manner.

Inspired by total-variation-regularized denoising for 2D images, the optimization-based approach usually introduces a smoothness prior as a regularization term in an optimization problem. The general formulation is
\begin{equation*}
\arg \min_{\widehat{\X}} \left\| \widehat{\X} - \X_{\epsilon}  \right\|_F^2 + \lambda J(\widehat{\X}),
\end{equation*}
where $J(\widehat{\X})$ is a regularization function, which can be specified in various ways. For example,~\cite{LozesEL:15} considers partial-differential-equation-based regularization and
~\cite{ZengCNPY:18} considers graph-Laplacian-based regularization.

Some deep-neural-network-based approaches are proposed recently to improve the denoising performance~\cite{DuanCK:19, Pistilli:19}. For example, neural projection denoising (NPD)~\cite{DuanCK:19} is a two-stage denoising algorithm, which first estimates reference planes and follows by projecting noisy points to estimated reference planes. It is a deep-neural-network version of weighted multi-projection~\cite{duanMWP}. NPD uses PointNet-based backbone to estimate a reference planes for each 3D point in a noisy point cloud; and then, projects noisy 3D points onto estimated reference planes and obtain denoised 3D points.

\subsubsection{3D point cloud downsampling}  
\begin{figure}
  \begin{center}
\includegraphics[width=0.9\columnwidth]{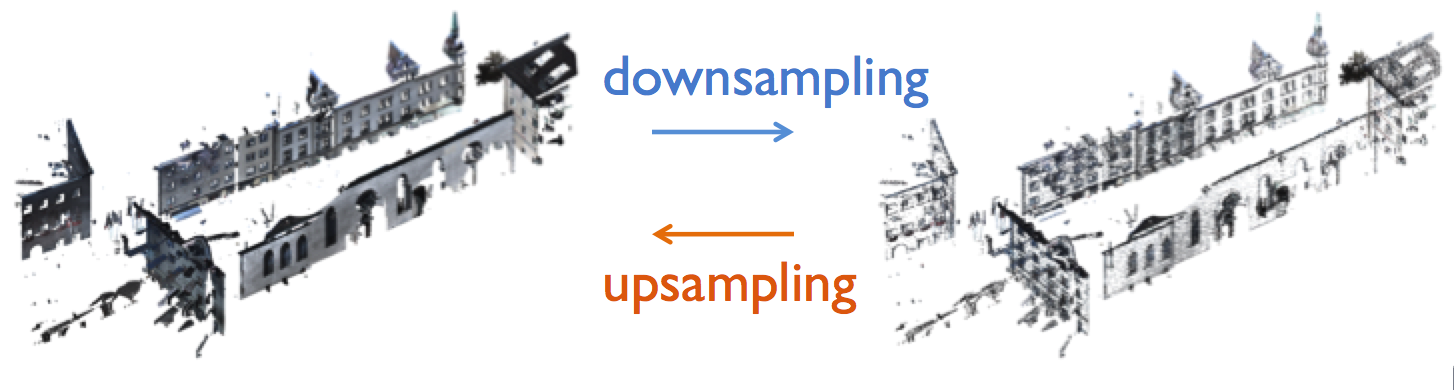}  
  \end{center}
  \caption{\label{fig:sampling} 
  3D point cloud downsampling and upsampling are primal and dual tasks, where downsampling aims to select a subset of 3D points in an original 3D point cloud while peserving representative information; and upsampling aims to generate a dense (high-resolution) 3D point cloud from a sparse (low-resolution) 3D point cloud to describe the underlying geometry of an object or a scene. }
\end{figure}

The goal of downsampling is to select a subset of 3D points in an original 3D point cloud while peserving representative information; see Figure~\ref{fig:sampling}. Handling a large number of 3D points is challenging and expensive. Therefore, a 3D point cloud is often sampled to a size that can be processed more easily. As a typical task of 3D point cloud processing, downsampling is potentially useful to data storage and the map creation module in autonomous driving. To represent a 3D scene, one can select representative 3D points from an HD map through downsampling, leading to faster and better localization performances~\cite{ChenTFVK:18}.

Let $\S = \{\x_i \in \R^d \}_{i=1}^N$ be a 3D point cloud with $N$ 3D points and $h$ be a downsampling operator that selects $M$ 3D points from $\S$, where $M < N$. The downsampling process works as
\begin{equation*}
  {\S}_\M = h(\S),
\end{equation*}
where ${\S}_\M = \{\x_{\M_i} \in \R^d \}_{i=1}^M$ is a downsampled 3D point cloud, where the downsampled set
$\M = (\M_1, \ldots , \M_{M})$ denotes the sequence of downsampled indices, $\M_i \in \{1, \ldots, N\}$ with $|\M| = M$.

\mypar{Standard experimental setup}
As a processing task, it is difficult to directly evaluate the performance of downsampling. Researchers usually input downsampled 3D point clouds to some subsequent tasks and test their performance.~\cite{ChenTFVK:18} evaluates downsampling on 3D point cloud registration. The evaluation metric is the localization error, such as the mean square error; and~\cite{DovratLA:18} suggests evaluating downsampling on the tasks of classification and reconstruction. For classification, the dataset is ModelNet40~\cite{WuSKYZTX:15} and the evaluation metric is classification accuracy. For reconstruction, the dataset is ShapeNet~\cite{ChangFGHHLSSSSX:15}  and  the evaluation metric is the reconstruction error, such as the Earth mover's distance~\eqref{eq:emd} and Chamfer distance~\eqref{eq:chamfer}.

\mypar{Standard methods}
There are three common approaches: farthest point sampling, learning-based sampling and nonuniformly random sampling.

A simple and popular downsampling technique is the farthest point sampling (FPS)~\cite{QiYSG:17}. It randomly chooses the first 3D point and then iteratively chooses the next 3D point that has the largest distance to all the points in the downsampled set. It is nothing but the deterministic version of K-means++~\cite{ArthurV:07}. Compared with uniformly random sampling, it has better coverage of the entire 3D point point given the same number of samples; however, FPS is agnostic to a subsequent application, such as localization and recognition. 

S-NET~\cite{DovratLA:18} is a deep-neural-network-based downsampling system. It takes a 3D point cloud and produces a downsampled 3D point cloud that is optimized for a subsequent task. The architecture is similar to latentGAN used for 3D point cloud reconstruction~\cite{AchlioptasDMG:18}. The difference is that S-NET does not reconstruct all the 3D points, but only reconstruct a fixed number of 3D points.  The loss function include a reconstruction loss, such as the Earth mover's distance~\eqref{eq:emd} and Chamfer distance~\eqref{eq:chamfer}, and a task-specific loss, such as classification loss. Since the reconstructed 3D point cloud is not  a subset of the original 3D point cloud any more, S-NET matches each reconstructed 3D point to its nearest neighbor in the original 3D point cloud; however, it is not trivial to apply S-NET to train and operate on large-scale 3D point clouds, which makes it less practical in autonomous driving.

To make the downsampling process more efficient and
adaptive to subsequent tasks,~\cite{ChenTFVK:18} considers a randomized downsampling strategy by choosing downsampled indices from a nonuniform distribution. Let $\pi \in \R^N$ be a downsampling distribution, where $\pi_i$ denotes the probability of
selecting the $i$th sample in each random trial.~\cite{ChenTFVK:18} designs an optimal downsampling distribution by solving a reconstruction-based optimization problem. It turns out that the optimal downsampling distribution is
$
    \pi^*_i  \ \propto \  \left\|\Hh_i\right\|_2,
$
where $\Hh_i \in \R^D$ is task-specific features of the $i$th point, which could be obtained by graph filtering.

\subsubsection{3D point cloud upsampling}
The goal of upsampling is to generate a dense (high-resolution) 3D point cloud from a sparse (low-resolution) 3D point cloud to describe the underlying geometry of an object or a scene. 3D point cloud upsampling is similar in nature to the super resolution of 2D images and is essentially an inverse procedure of downsampling; see Figure~\ref{fig:sampling}.   It is potentially useful to reduce the cost by making use of a low-resolution LiDAR in autonomous driving.

Let $\S = \{\x_i \in \R^d \}_{i=1}^N$ be a 3D point cloud with $N$ 3D points and $h$ be a upsampling operator that generate $N'$ 3D points from $\S$, where $N' > N$. The upsampling process works as
\begin{equation*}
  \widehat{\S} = h(\S),
\end{equation*}
where $\S \subset \widehat{\S}$. Intuitively, a 3D point cloud $\S$ is sampled from some surface and we aim to use the high-resolution 3D point cloud $\widehat{\S}$ to capture the same surface, but provide a higher density.

\mypar{Standard experimental setup} 
There is no standard benchmark for 3D point cloud upsampling. Researchers create their own training and testing datasets based on the VisionAir repository~\cite{Visionair:17}, ModelNet40~\cite{WuSKYZTX:15}, ShapeNet~\cite{ChangFGHHLSSSSX:15}, or SHREC15~\cite{LianZCEEFGGLLLL:15}. Some common evaluation metrics are the Earth mover's distance~\eqref{eq:emd} and Chamfer distance~\eqref{eq:chamfer}.

\mypar{Standard methods} 
Classical 3D point cloud upsampling algorithms are based on image super resolution algorithms. For example,~\cite{AlexaBCFLS:03} constructs surfaces with the moving least squares algorithm and generates new points at the vertices of the Voronoi diagram to upsample a 3D point cloud; to avoid over-smoothing,~\cite{HuangWGCAZ:13} applies an anisotropic locally optimal projection operator to preserve sharp edges by pushing 3D points away from the edges, and achieves the edge-aware 3D point cloud upsampling;~\cite{WuHGZC:15} combines the smoothness of surfaces and the sharpness of edges through an extracted meso-skeleton. The meso-skeleton consists of a mixture of skeletal curves and sheets to parameterize the underlying surfaces. It then generates new 3D points by jointly optimizing both the surface and 3D points residing on the meso-skeleton; however, these classical upsampling algorithms usually depend heavily on local geometry priors, such as the normal vectors and the curvatures. Some algorithms also suffer from multiscale structure preservation due to the assumption of global smoothness~\cite{WangWHCS:18}.

With the development of deep neural networks, more upsampling algorithms adopt the learning-based approach. PU-Net~\cite{YuLFCH:18a} is the first end-to-end 3D point cloud upsampling network, which extracts multi-scale features based on PointNet++. The architecture is similar to latentGAN for 3D point cloud reconstruction, but reconstructs many more 3D points than the original 3D point cloud. The loss function includes a reconstruction loss and a repulsion loss, pushing a more uniform distribution for the generated points. Inspired by the recent success of neural-network-based image super-resolution, patch-based progressive~\cite{WangWHCS:18} proposes a patch-based progressive upsampling architecture for 3D point clouds. The multi-step upsampling strategy breaks an upsampling network into several subnetworks, where each subnetwork focuses on a specific level of details. To emphasize edge preservation, EC-Net designs a novel edge-aware loss function~\cite{YuLFCH:18b}. During the reconstruction, EC-Net is able to attend to the sharp edges and provide more precise 3D reconstructions. Note that all those deep-neural-network-based methods are trained based on well-selected patches, which cover a rich variety of shapes.

\subsubsection{3D point cloud registration}
\label{sec:registration}
The goal of registration is to transform multiple 3D point clouds from local sensor frames into the standardized global  frame. The key idea is to identify corresponding 3D points  across frames and find a transformation that minimizes the distance (alignment error) between those correspondences.
As a typical task of 3D point cloud processing, 3D point cloud registration is critical to the map creation module and the localization module of autonomous driving. In the map creation module, we need to register multiple LiDAR sweeps into the standardized global frame, obtaining a point-cloud map. In the localization module, we need to register a real-time LiDAR sweep to the point-cloud map to obtain the pose of the autonomous vehicle, which includes the position and the heading.

Let $\mathbf{S}=\{\S_i\}_{i=1}^K$ be $K (\geq 2$) frames of observed 3D point clouds, where the $i$th frame $\S_i \in \R^{N_i \times 3}$ is a 3D point cloud with $n_i$ points. Let $\mathbf{P}=\{p_i\}_{i=1}^K$ be the corresponding sensor poses, where the $i$th sensor pose $p_i \in SE(3)$ is a 3D Euclidean rigid transformation. The registration process aims to estimate the sensor pose of each frame by solving the following optimization problem,
\begin{equation}
\label{eq:registration}
    \widehat{\mathbf{P}} \ = \ \operatorname*{arg\,min}_{\mathbf{P}} \mathcal{L}_{\mathbf{S}} (\mathbf{P}),
\end{equation}
where $\widehat{\mathbf{P}}$ is the final estimated sensor poses and $\mathcal{L}_{\mathbf{S}} (\mathbf{P})$ is the loss function parameterized by the poses $\mathbf{P}$. It evaluates the registration quality using the correspondences under the estimated poses.


\mypar{Standard experimental setup}
For large-scale outdoor 3D point cloud registration, there are several standard datasets, including the KITTI dataset~\cite{GeigerLU:12}, the Oxford RobotCar dataset~\cite{RobotCarDatasetIJRR}, and the ETH ASL dataset~\cite{Pomerleau2012}. To evaluate the registration performance of a pair of 3D point clouds, one can simply calculate the position and orientation differences between the estimated pose and the ground truth. To evaluate on a sequence of 3D point clouds, absolute trajectory error (ATE)~\cite{sturm2012benchmark} is a commonly used metric. It is the $\ell_2$ norm of the positional residual vector after aligning the estimated trajectory with the ground truth through a global rigid transform.

\mypar{Standard methods} 
A point cloud registration method usually solve the optimization~\eqref{eq:registration} by iteratively alternating the correspondence search and alignment, especially when the correspondence search is affected by the current alignment. Once the alignment errors fall below a given threshold, the registration is said to be complete. Classical point cloud registration methods can be roughly group into the following categories: pairwise local registration, pairwise global registration, and multiple registration. Pairwise registration deals with only two adjacent frames, where $K=2$. The local registration methods assume a coarse initial alignment between two point clouds and iteratively update the transformation to refine the registration. For example, the iterative closest point (ICP) algorithms~\cite{BeslM:92,chen1992object,Rusinkiewicz_EffiICP_DDIM01} and the probabilistic-based algorithms~\cite{Jian_RegMixGaussians_ICCV05,Myronenko_CPD_PAMI10,Danelljan_ProbColorPCReg_CVPR16} fall into this category. The local registration methods are well-known for requiring a ``warm start'', or a good initialization, due to limited convergence range.
The global registration methods~\cite{Yang_GoICP_PAMI16, Aiger_4PCS_ACMG08,Mellado_Super4PCS_CGF14,Zhou_FastGlobal_ECCV16} do not rely on the ``warm start'' and can be performed on 3D point clouds with arbitrary initial poses. Robust estimations, such as RANSAC~\cite{Fischler_RANSAC_CACM81}, are typically applied to handle the incorrect correspondences. Most global methods extract feature descriptors from two 3D point clouds, which establish 3D-to-3D correspondences for relative pose estimation. These feature descriptors are either hand-crafted features, such as FPFH~\cite{rusu2009fast}, SHOT~\cite{tombari2010unique}, PFH~\cite{rusu2008aligning}, spin-images~\cite{johnson1999using}, or learning-based features, such as 3DMatch~\cite{zeng20173dmatch}, PPFNet\cite{deng2018ppfnet}, and 3DFeat-Net\cite{yew20183dfeat}.
In addition to pairwise registration, several multiple registration methods~\cite{Theiler_GloTerrGraOpt_JPRS15,Evangelidis_JRMPC_ECCV14,Izadi_KinectFusion_ACMUIST11,Torsello_MultiRegDualQuant_CVPR11,Choi_ReconIndoor_CVPR15} have been proposed, which incrementally add a coming 3D point cloud to the model registered based on all the previous 3D point clouds. The drawback of the incremental registration is the accumulated registration error. This drift can be mitigated by minimizing a global cost function over a graph of all sensor poses~\cite{Theiler_GloTerrGraOpt_JPRS15,Choi_ReconIndoor_CVPR15}.

Recent works explore deep-neural-network-based approaches to solve registration problems~\cite{Ding_2019_CVPR,LuWZFYS:19,Henriques_MapNet_CVPR18,CodeSLAM_Bloesch_2018_CVPR}. Some image registration methods use unsupervised learning to exploit inherent relationships between depth and motion. This idea is further explored in~\cite{CodeSLAM_Bloesch_2018_CVPR, Zhou_DeepTAM_ECCV18, Yand_DVSO_ECCV18,Li_DL2DLoopClos_IROS17} using deep learning for visual odometry and SLAM problems. Methods in~\cite{Henriques_MapNet_CVPR18,Parisotto_GlbPoseEstAttRNN_CVPR18} use the recurrent neural network (RNN) to model the environment through a sequence of depth images in a supervised setting. For example, MapNet~\cite{Henriques_MapNet_CVPR18} develops a RNN for RGB-D SLAM problem where the
registration of camera sensor is performed using deep template matching on the discretized spatial domain. Unlike other learning-based methods, DeepMapping~\cite{Ding_2019_CVPR} uses deep neural networks as auxiliary functions in the registration optimization problem, and solves it by training those networks in an unsupervised way. The learnable free-space consistency loss proposed in DeepMapping allows it to achieve better performances than ICP and its variants.

\bibliographystyle{IEEEbib}
\bibliography{refs}
\end{document}